\documentclass{article}

\PassOptionsToPackage{numbers, compress}{natbib}

\usepackage[preprint]{neurips_2026}

\usepackage[utf8]{inputenc} 
\usepackage[T1]{fontenc}    
\usepackage{hyperref}       
\usepackage{url}            
\usepackage{booktabs}       
\usepackage{amsfonts}       
\usepackage{nicefrac}       
\usepackage{microtype}      
\usepackage{xcolor}         

\usepackage{amsmath,amssymb,amsthm}
\usepackage{graphicx}
\usepackage{caption}
\usepackage{subcaption}
\usepackage{float}
\title{Bifurcations from Weight Space: JEPAWG on Parametric Dynamical Systems}

\author{%
  Tobias G\"{o}bel$^{1,2}$ \\
  \And
  Miranda C. N. Cheng$^{1,3}$\thanks{Correspondence to: Miranda C. N. Cheng \texttt{<miranda.cheng.w@gmail.com>}.} \\
  \AND
  \mdseries $^{1}$Institute for Mathematics, Academia Sinica, Taipei, Taiwan \\
  $^{2}$Department of Physics, National Taiwan University, Taipei, Taiwan \\
  $^{3}$Institute of Physics, University of Amsterdam, Amsterdam, Netherlands \\
}

\begin{document}
\raggedbottom
\maketitle

\begin{abstract}
Trained network weights are increasingly treated as data, but there is rarely ground truth about what a population of models should encode, which hampers systematic validation of weight-space methods. This makes parametric dynamical systems a natural test-bed: the dynamics change qualitatively at submanifolds of parameter space that are, in well-studied families, known analytically or numerically. Fine-tuning a forecasting network at each point of a parameter grid yields a model zoo indexed by a physical parameter, against which weight-space diagnostics can be checked.

We build such model zoos for four dynamical systems and analyse them with JEPAWG, a recently proposed joint-embedding hypernetwork built around a learned parameter--weight representation. It was first introduced for neural samplers for lattice field theories, where its latent geometry located the phase transition and estimated the 2D Ising critical exponent. 
Applied to dynamical systems, JEPAWG is competitive on forecasting with a hypernetwork trained directly for that task, despite being supervised on the parameter--weight correspondence rather than on forecasts. Beyond forecasting, the norm of the parameter-derivative of the decoded weights locates Hopf points and curves, tracks a period-doubling cascade closely enough to give a qualitative estimate of the Feigenbaum constant, and detects a global bifurcation, which has no eigenvalue signature and is therefore invisible to linearisation. As such, JEPAWG provides interpretable insight into the dynamics that is not available from other hypernetworks. Together, these results indicate that JEPAWG is a workable instrument for reading physical
structure off a population of trained networks.
\end{abstract}
\begin{NoHyper}
{\renewcommand{\thefootnote}{}%
 \footnotetext{A version of this work appears at the NeurIPS 2026 Workshop on Neural Network Artifacts as a New Data Modality.}}
\end{NoHyper}
\section{Introduction}
\label{sec:intro}

With the growing number and scope of trained neural networks, the
weights have increasingly become an object of study themselves. Collections of networks trained under
controlled variation, \emph{model zoos}~\citep{schürholt2022modelzoosdatasetdiverse}, are
used as datasets in their own right: to predict properties of a
network from its weights, to average models into stronger ones
(model soups~\citep{wortsman2022modelsoupsaveragingweights}), or to learn representations of
weight space itself. A related line of work builds \emph{hypernetworks}~\citep{ha2016hypernetworks} generating
the weights of a target network from a conditioning input, treating weights not as
data but as the output to be produced.

A model zoo becomes a \emph{parametrised} model zoo when each of its
members is associated with a value of a continuous task parameter $p$. Because the task the members are trained for differs only through
$p$, such a zoo carries information about the system the task operates in, and analysing the zoo, rather than any single network in
it, becomes a way to study that system. JEPAWG (JEPA Weight Generation)~\citep{jepawg} realises this vision through an interpretable
hypernetwork: it learns a joint latent space of model weights and
task parameters with a joint-embedding predictive architecture (JEPA) objective~\citep{assran2023ijepa}. Afterwards, it
trains a decoder from this latent back to weight space, which enables the generation of weights at unseen parameters. The parameter derivative of the decoded weights exposes qualitative changes in the underlying
dynamics. JEPAWG was first applied in lattice field theory, where its latent space was shown to locate the Ising
phase transition and estimate its critical exponent. Here we consider
parametric dynamical systems, which provide a natural test-bed:
data generation is cheap, and the dynamics change qualitatively on
bifurcations, submanifolds of parameter space whose location is, for
the families we study, known analytically or numerically.

In this work we build parametrised model zoos for four dynamical
systems, train the JEPAWG architecture, and study the result from two perspectives: as a hypernetwork, by evaluating the generated forecasters at
unseen parameters, and as a probe of the dynamics, by testing
whether the derivative of the decoded weights with respect to the
control parameters localises the known bifurcations. We find that JEPAWG is competitive with
an end-to-end hypernetwork on held-out interpolation
parameters, despite never being trained on the forecasting loss. As a probe for the underlying dynamical systems, the
weight-space response localises Hopf points and curves, resolves a
period-doubling cascade, and registers a global bifurcation. Our main contributions are the following.

\paragraph{A weight-space benchmark with known ground truth}
Parametric dynamical systems supply an independent statement of what a
population should encode. Crossing a bifurcation qualitatively changes the
dynamics, so networks trained on either side should differ in weight space.
We construct zoos for four families with established bifurcation structure,
against which weight-space diagnostics can be validated: the Brusselator Hopf
curve $b_c = 1 + a^2$, the Lorenz Hopf locus $\rho_H(\sigma)$ at $\beta = 8/3$, the period-doubling cascades of the R\"ossler oscillator and the H\'enon map, whose
accumulation should recover the Feigenbaum constant $\delta \approx 4.6692$.
 
\paragraph{Detection of local bifurcations}
The parameter derivative of the decoded weights peaks where the dynamics
are most sensitive to the parameter, which happens near local
bifurcations. For the Lorenz system ($\sigma = 10$, $\beta = 8/3$) the peak
lies at $\rho = 24.51 \pm 0.23$, inside the bistable window preceding the
subcritical Hopf point $\rho_H \approx 24.74$, and tracks $\rho_H(\sigma)$ when
both parameters are swept. The response also locates the supercritical
Hopf curve of the Brusselator, and in the H\'enon map it resolves the
first five period doublings accurately enough to provide a qualitative estimate of the Feigenbaum constant $\delta$.

\paragraph{Detection of a global bifurcation} Global bifurcations, unlike local bifurcations, are not visible in the linearisation and traditionally require computing invariant manifolds or scanning attractor-level observables. The same derivative again detects the Lorenz Bykov T-point, at which every fixed point retains its stability. This shows that information inaccessible to a Jacobian computation can be read off the geometry of the JEPAWG-generated weights.

\paragraph{The decoded map as a regulariser of the weight grid} The
fine-tuned weights carry per-run training noise that dominates their
parameter derivative. The decoder, fitted on a fixed embedding of the parameter, does not
reproduce this noise (Fig.~\ref{fig:lorenz-bif}), and the decoded
weights retain accurate forecasts for longer than the fine-tuned
weights on all four systems (Table~\ref{tab:raw-vs-decoded}). The JEPA-style latent is organised by dynamical regime, has an intrinsic dimension that is close to $\dim p$ (Fig.~\ref{fig:brusselator-latent}, Table~\ref{tab:embed}), and is insensitive to parameters that do not enter the dynamics (Appendix~\ref{app:embedding}, Fig.~\ref{fig:randomp}).

\section{Related Work}
\label{sec:related}
\textbf{Trained weights as data.} Model zoos
\cite{schürholt2022modelzoosdatasetdiverse} assemble populations of networks
trained under controlled variation, and permutation-aware encoders and neural
functionals \cite{navon2023equivariant, zhou2023neuralfunctionals} learn
structure over weight space directly. More recently \citet{modelhubs} argues
that useful weight-space representations can be learned from unstructured
public model repositories, without curated zoos. That line targets prediction
of model properties across heterogeneous populations. We study a single population indexed by a physical parameter, where known
bifurcation structure provides ground truth for weight-space diagnostics.

\textbf{Hypernetworks for dynamical systems.} Hypernetworks, networks that
generate the weights of another network \cite{ha2016hypernetworks}, have
recently been applied to parametric dynamical systems. PHLieNet
\cite{vlachas2026phlienet} learns a continuous parameter-to-weights map by
back-propagating through the forecasting loss, generalising recurrent
forecasters that take $p$ as an additional input. It is indexed by the
control parameter, but its weights are constrained
only by forecasting accuracy, and in our experiments its parameter derivative
shows no discernible bifurcation signal (Section~\ref{sec:interp-lorenz}).
CoDA \cite{kirchmeyer2022generalizingnewphysicalsystems} adapts a shared forecaster through
context-dependent low-rank weight corrections, and TiDHy \cite{tidhy}
generates, from a latent state, the weights that mix several linear systems
evolving on different timescales. Both condition on a variable inferred
from trajectories rather than on the control parameter, so their weights
are not indexed by a directly interpretable physical quantity. In each case the weights are a means to accurate prediction. In this work forecasting remains a requirement, but the map from parameter to weights carries the additional role of a diagnostic.

\textbf{Transitions read off learned models.} Machine learning has been used
to locate phase transitions without a known order parameter, from
classifiers trained on raw configurations \cite{Carrasquilla_2017} to the
accuracy of classifiers trained on deliberately mislabelled data
\cite{vannieuwenburg2017learning}. Closer to the present work are approaches
that read the transition off the parameters of trained networks rather than
their outputs. \citet{Tanaka_2017} observed that the weights of a
network trained to predict the temperature of Ising configurations organise
into two domains whose boundary approximates the critical temperature. \citet{nqsweights} fine-tune neural quantum states adiabatically across a
phase diagram, so that the resulting weights are strongly correlated, and
detect quantum phase transitions as structure in weight space. JEPAWG
\cite{jepawg} instead learns a map from couplings to the weights of
normalising flows for a scalar lattice field theory through a JEPA-style
\cite{assran2023ijepa} latent space, whose geometry locates the phase
transition and encodes a finite-size shift consistent with the 2D Ising
critical exponent. In the present work we transfer this approach to
dynamical systems, where the known bifurcation structure provides ground
truth.

Within dynamical systems, data-driven approaches to bifurcations have
stayed close to local, equilibrium-based theory. \citet{bury2021deep,
Bury_2023} train classifiers on noisy normal-form simulations to
detect the critical slowing down that precedes a bifurcation in a time
series and predict its type, including period doubling in discrete-time
maps. \citet{mogharabin2025bifurcation} fit a normal form on a learned
low-dimensional invariant manifold, and
\citet{pal2026machinelearningcomplexsystems} locate saddle-nodes from a
learned equilibrium-to-parameter map trained on the governing equations.
All of these target bifurcations of a single equilibrium, and
\citet{Bury_2023} explicitly leave global bifurcations outside their
scope. We instead differentiate a map from the parameter to the weights of
a model of the full dynamics. This requires no order parameter,
linearisation or governing equations, and remains defined for transitions
that are not a property of any single equilibrium or periodic orbit, such
as global bifurcations \cite{glendinning1986,Barrio2012KneadingsSD}.
\section{Background}

\label{sec:background}

\paragraph{Parametric dynamical systems and bifurcations}
\label{sec:setup}
Consider a parametric dynamical system whose evolution is governed by
\begin{equation}
    \frac{dx}{dt} \;=\; f(x;\,p),
    \label{eq:cont}
\end{equation}
where $x\in\mathbb{R}^{D_x}$ is the state, $t$ is time, and
$p\in\mathbb{R}^{D_p}$ is a vector of control parameters. Trajectories
can be obtained with any integration scheme (e.g.\ Euler or
Runge--Kutta). For discrete-time maps the evolution is given by
$x_{n+1}=F(x_n;p)$, producing a discrete sequence of states instead of a continuous path.

The long-time behaviour of such a system is characterised by its
\emph{attractors}, the sets toward which trajectories converge. Depending on the parameter $p$, different types of attractors can arise: fixed
points, limit cycles (periodic orbits), or strange attractors, on
which the dynamics are chaotic. As the control parameters vary, an attractor can lose stability
and the dynamics reorganise qualitatively; such a change is called a
\emph{bifurcation}. The submanifolds of parameter space on which
bifurcations occur partition it into regions of qualitatively distinct
dynamics \citep{strogatz:2000, guckenheimer1983nonlinear}. Two types recur in this work: the \emph{Hopf
bifurcation}, in which a fixed point loses stability, and the
\emph{period-doubling bifurcation}, in which a periodic orbit of
period $\tau$ is replaced by one of period $2\tau$. Many systems reach
chaos through an infinite cascade of such doublings.

\emph{Local} bifurcations, including the two types mentioned above, happen at a single fixed point
or orbit and are visible in the linearisation there: a Hopf
bifurcation is an eigenvalue pair of the Jacobian crossing the
imaginary axis, a period doubling a Floquet multiplier crossing $-1$.
Such conditions can be solved in closed form or tracked numerically,
which is how the analytic reference values we validate against in this work
are obtained \citep{guckenheimer1983nonlinear}.

\emph{Global} bifurcations admit no such local criterion: there is no
single fixed point or orbit whose linearisation registers the change.
They instead involve the invariant manifolds of a saddle. A saddle
carries an \emph{unstable manifold}, the trajectories that emerge from
it, and a \emph{stable manifold}, the trajectories that converge to
it. Both extend far into state space. A one-dimensional unstable
manifold generically misses the stable manifold of a saddle, since
their coincidence is one scalar condition on the parameters; at
isolated parameter values, however, it lands exactly on one, either of the
same saddle (a \emph{homoclinic} connection) or of another (a
\emph{heteroclinic} connection). When this happens the global
arrangement of trajectories reorganises, yet every fixed point keeps
its stability. As a result there are no eigenvalue signals for this
change. Finding global bifurcations therefore traditionally requires
computing the invariant manifolds themselves, or scanning
attractor-level observables across parameter space.

In this work we locate both classes by a route that uses neither the
linearisation nor an explicit analysis of trajectories. At each
parameter value $p_i$ we train a small network ($\sim 10^4$ weights) to forecast the dynamics, so
that the model zoo supplies a family of trained weight vectors
$\{\theta(p_i)\}$. 
Each network is fine-tuned only on trajectories generated at its own $p_i$, so its weights encode how the system behaves there; when the dynamics reorganise, the weights must change rapidly. A bifurcation is then visible as a large sensitivity of the weights to the parameter, a quantity we make precise in Section~\ref{sec:proxy}.

The first step is to build a forecasting model, by which we mean a function that predicts the future evolution of the state from a short observed history together with the parameter values $p$. We call such a function a \emph{parametric forecaster}, and specify the
architecture we use below.

\paragraph{Forecasting}
\label{sec:forecasting-networks}
The forecasting task is to predict the evolution of a state $x_t$ given its history $(x_{t-T+1},\ldots,x_t)$
and the parameter vector $p$. 
The next state is obtained through a neural network that predicts the finite difference:
\begin{equation}
    \hat{d}_t \;=\; f_{\theta}\bigl(x_{t-T+1:t};\,p\bigr),
    \qquad
    x_{t+1} \;\approx\; x_t + \Delta t\cdot\hat{d}_t,
    \label{eq:derivative-matching}
\end{equation}
where $T$ is the input sequence length (the history-window size), $\Delta t$ is the observation interval ($\Delta t = 1$ for discrete maps) and
$\theta$ are the network weights.
Given windows drawn from the trajectories at each parameter, the weights
are fitted by minimising the derivative-matching loss
\begin{equation}
    \mathcal{L}_p(\theta)
    \;=\;
    \sum_{k,t}\bigl\|f_{\theta}(x_{t-T+1:t}^{(k)};\,p)
    - (x_{t+1}^{(k)} - x_t^{(k)})/\Delta t\bigr\|^2,
    \label{eq:dm-loss}
\end{equation}
where $k,t$ index initial conditions and window positions.
The forecasting networks considered in this work are trained with this objective: for the $p$-conditional network of
Section~\ref{sec:datagen} (summed over the training parameters $p_i$); for the per-parameter fine-tuning that produces
each $\theta_i$; and for the hypernetwork baseline below, where $\theta$
is generated from $p$.
Models that have been employed for learning this type of sequential data include Recurrent Neural Networks (RNNs) \cite{graves2014generatingsequencesrecurrentneural}, LSTMs \cite{6795963,vlachas2020backpropagationalgorithmsreservoircomputing}, Transformers \cite{NIPS2017_3f5ee243, duth2026mechanisticanalysistransformersdynamical} and Temporal Convolutional Neural Networks with Causal Dilated convolutions (TCNN-CD) \cite{bai2018empiricalevaluationgenericconvolutional}. Throughout this work we use the TCNN-CD architecture as the forecasting
network $f_\theta$, whose stacked causal convolutions, with exponentially
growing dilations, provide a long receptive field without recurrence.

\subsection{Forecasting Hypernetworks}
\label{sec:phlienet}
A forecaster can be made aware of the parameter vector $p$ in several ways.
The simplest is not to condition on $p$ at all and to let the model infer the dynamics from the state history alone. Alternatively, extending the state vector as $x^\prime = [x, p]$ gives the model direct knowledge of the parameter setting. As a third option, we consider a hypernetwork: a separate neural network that learns to map a given parameter to a weight vector, $p \mapsto \theta$. The resulting weights $\theta$ then parametrise the forecasting network $f_\theta$, which predicts the state evolution.

\paragraph{PHLieNet}
One such hypernetwork for dynamical systems is PHLieNet~\citep{vlachas2026phlienet},
which is trained end-to-end. A Learned Interpolated Embedding (LIE) maps the
parameter $p$ to a continuous embedding $e(p)\in\mathbb{R}^{d_e}$ by
softmax-RBF interpolation between learned embedding vectors placed at fixed
anchor positions, and a small MLP, the hypernetwork $\mathrm{HN}$, maps $e(p)$
to the flattened weight vector $\theta$ of the TCNN-CD. Only the anchor
embeddings and the hypernetwork are trainable; the TCNN-CD weights are
\emph{generated} per parameter. Training minimises the loss
\eqref{eq:dm-loss} at the generated weights, summed over the set of parameters,
$\sum_i \mathcal{L}_{p_i}\bigl(\mathrm{HN}(e(p_i))\bigr)$.

Two consequences of this design matter for what follows. First, the
parameter-to-weights map is supervised \emph{only} through the
forecasting loss, which leaves the weight-space geometry unconstrained: the map is free to organise the latent as long as the forecasts are accurate. Second, the LIE anchors only cover
the training range by construction and softmax-RBF
interpolation saturates outside the anchor support, so PHLieNet is an
interpolation architecture by design. 
 
\paragraph{JEPAWG}
\label{sec:jepawg}
Instead of training a hypernetwork through the forecasting loss, one can
treat trained network weights themselves as data. This is the approach of
JEPAWG~\cite{jepawg}. First, one prepares a collection of per-parameter
weights $\{(p_i,\theta_i)\}_{i=1}^M$ by training a forecaster at each
parameter value. A joint-embedding predictive architecture
(JEPA)~\cite{assran2023ijepa} is then trained on that collection: the
parameters and the weights are treated as two views of the same object,
and the weight embedding is predicted from the parameter embedding, so
that the objective is defined in latent space. A decoder back to weight
space is trained separately, and weights at a new parameter are obtained
by composing the parameter encoder, the predictor and the decoder.
Concretely, JEPAWG has four learnable components:
a weight encoder $E_\theta$, a parameter encoder $E_p$, a latent
predictor $P$, and a decoder $D$. The first three are trained jointly, and the decoder
afterwards with them frozen:
\begin{description}
\item[Stage 1] Train $(E_\theta,E_p,P)$ with the JEPA/VICReg~\cite{bardes2022vicreg} loss
\begin{equation}
    \mathcal{L}_{\rm JEPA}
    = w_{\text{inv}}\|P(z_c)-z_w\|^2
    + w_{\text{var}}\bigl(\text{V}(z_w)+\text{V}(z_c)\bigr),
    \qquad
    z_w = E_\theta(\theta),\ z_c = E_p(p).
    \label{eq:stage1}
\end{equation}
The variance term
$\text{V}(z)=\frac{1}{d_z}\sum_{j=1}^{d_z}\max\bigl(0,\gamma-s_j(z)\bigr)$,
with $s_j(z)$ the standard deviation of latent dimension $j$ over
the batch and margin $\gamma=1$, prevents the encoders from collapsing
to a constant. The standard VICReg covariance term is omitted
($w_{\text{cov}}=0$).
\item[Stage 2] Freeze $E_\theta, E_p, P$ and train the decoder by MSE in weight space:
\begin{equation}
    \mathcal{L}_{\rm Dec}
    \;=\;
    \bigl\|D\bigl(P(E_p(p_i))\bigr) - \theta_i\bigr\|^2.
    \label{eq:stage2}
\end{equation}
\end{description}
At inference, weights are generated by composing the parameter encoder,
latent predictor and decoder: $\hat{\theta}(p) = D\bigl(P(E_p(p))\bigr)$,
yielding forecasting networks at unseen parameters. Moreover, since the
latent space is organised by the parameter-weight correspondence, its
geometry can be probed for qualitative changes in the underlying
dynamics, such as bifurcations or phase transitions. 
\section{Methods}
\label{sec:methods}
This section describes the pipeline in three steps (Figure~\ref{fig:pipeline}).
Section~\ref{sec:datagen} builds the model zoos: trajectories are
generated across a parameter range, a single $p$-conditional forecaster is
trained on all of them, and that network is fine-tuned at each selected point
to give the weights $\{(p_i,\theta_i)\}_{i=1}^M$.
Section~\ref{sec:jepawg-training} fits JEPAWG to that collection of weights.
Section~\ref{sec:analysis} defines the two ways we evaluate the resulting
map $p\mapsto\hat\theta(p)$: as a forecaster at unseen parameters, and as
a probe of the dynamics through its parameter derivative.
\subsection{Data Generation}
\label{sec:datagen}
In principle, the zoo could be built by training a separate forecaster
from scratch at every parameter value. We instead train one
$p$-conditional TCNN-CD \cite{bai2018empiricalevaluationgenericconvolutional} on the pooled trajectories and use it as the initialisation for a further fine-tune at each $p_i$. Besides amortising the training cost, this addresses a symmetry of the problem. The loss is invariant under a large symmetry group, including permutations of channels within a layer and rescalings across ReLU layers, so forecasters trained independently at neighbouring parameter values would generically land in different symmetry copies of an equivalent solution, and the difference between their weight vectors would be dominated by that choice rather than by the change in the dynamics. A common initialisation acts as an approximate gauge fixing. The hope is that fine-tuning every $\theta_i$ from the common conditional
network, with a short schedule at a fixed learning rate, keeps the
weights within a common symmetry branch, so that variation along the grid
reflects the parameter dependence of the dynamics.
\paragraph{Trajectory generation}
For each system we sample $M$ parameter values by scrambled Sobol
sequences \cite{Sobol1967OnTD} over the control range (per-system ranges are given in Table~\ref{tab:all-inter-extrap}, and densities and
stratification in Table~\ref{tab:protocol}). At each parameter $p$ we generate
trajectories from several random initial conditions. Continuous-time systems
are integrated with fixed-step RK4 at solver step $\delta t = 10^{-3}$; the
observation interval, recorded length, and number of initial conditions are
per-experiment and are listed in Table~\ref{tab:protocol}. Discrete maps are iterated
directly ($\Delta t = 1$). To mitigate dependence on the initial conditions we discard an
initial transient for some systems, integrated in addition to the
recorded lengths above; durations
per system are listed in Table~\ref{tab:protocol}. States are
standardised per state dimension, with mean and variance computed
over the pooled training trajectories; training and rollouts operate
in this scaled space, and all metrics are computed after inverting
the standardisation, in the original state space.
\paragraph{Conditional forecaster}
\label{sec:conditional-training}
We first train a single $p$-conditional TCNN-CD on the trajectories from
all parameter values. The conditioning is implemented by appending $p$ to
every state, $x' = [x, p]$, and the network is trained by minimising
$\sum_i \mathcal{L}_{p_i}(\theta)$ with the loss \eqref{eq:dm-loss}. This
conditional base network learns the dynamics shared across the parameter
range and serves as a common initialisation for the per-parameter
fine-tuned networks below. Training windows are perturbed by additive Gaussian input noise with
standard deviation $5\%$ of the standardised feature scale, redrawn each batch and applied to the input
window only. Target states are computed from the clean trajectories, and
evaluation uses clean inputs. This trains the network to contract the
small state perturbations it feeds itself during iterated predictions,
following \cite{vlachas2026phlienet}.

\paragraph{Fine-tuning}
\label{sec:weight-grid}
Initialised from the conditional base, the network is fine-tuned on the
windows of each parameter value $p_i$ separately for $30$ epochs with Adam at learning
rate $10^{-3}$, under the same objective and input noise as
pre-training, giving target weights $\theta_i \in \mathbb{R}^{D_\theta}$. Each member keeps the standardised parameter as an extra input channel,
inherited from the conditional base; since it is fine-tuned at a fixed
$p_i$, this channel is constant during fine-tuning, and all
$p$-dependence the member learns is stored in $\theta_i$.
The resulting \emph{weight grid} $\{(p_i,\theta_i)\}_{i=1}^M$ is flattened and standardised per dimension.

\subsection{JEPAWG Training}
\label{sec:jepawg-training}
\begin{figure}[t]
    \centering
    \includegraphics[width=0.99\linewidth]{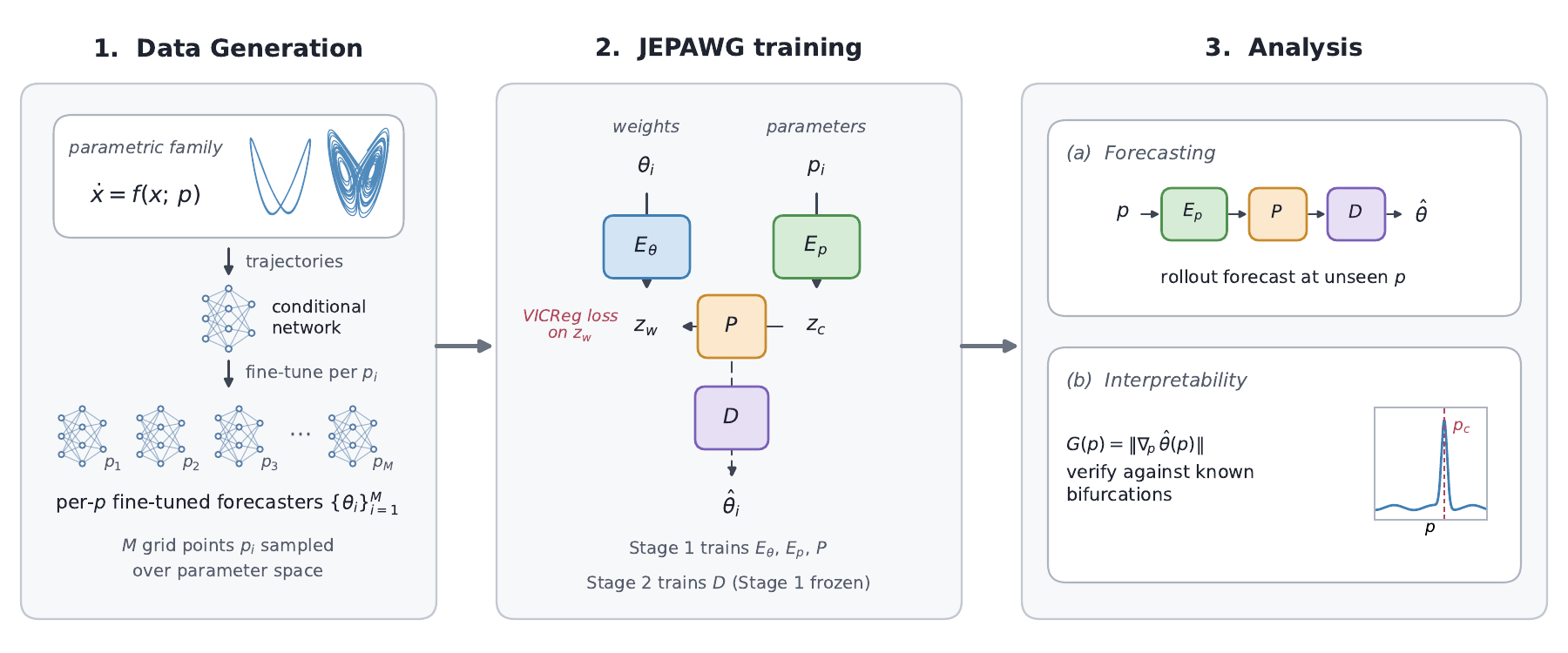}
    \caption{\textbf{JEPAWG pipeline.} \emph{(1)}~A conditional
forecaster is trained on trajectories from the parametric family and
fine-tuned at each of $M$ parameter values, yielding the weight grid
$\{(p_i,\theta_i)\}_{i=1}^{M}$ (Section~\ref{sec:weight-grid}).
\emph{(2)}~Encoders, predictor and decoder are trained in two stages
(Eqs.~\ref{eq:stage1},~\ref{eq:stage2}), giving the inference path
$\hat\theta(p)=D(P(E_p(p)))$. \emph{(3)}~The decoded weights are
evaluated as a hypernetwork and probed via
$G(p)=\|\nabla_{\!p}\hat\theta(p)\|$, verified against known
bifurcations.}
    \label{fig:pipeline}
\end{figure}

JEPAWG is then trained on the parameter--weight pairs
$\{(p_i,\theta_i)\}_{i=1}^{M}$ assembled above, in the two stages described in
Section~\ref{sec:jepawg} (Figure~\ref{fig:pipeline}, panel 2). All components $E_p, P, E_\theta, D$ are taken to be MLPs with ReLU activations. The latent
dimension is $d_z = 16$, following \cite{jepawg}: the latent is
deliberately over-sized relative to the dimension of the parameter
manifold, so that any concentration onto fewer dimensions is a learned
property rather than an architectural constraint.
Note that $E_p$ is not a compression: $p$ is one- or two-dimensional
here, so the parameter branch expands rather than reduces, and the name
\emph{encoder} is inherited from the joint-embedding setting rather than
descriptive.
The
weight encoder $E_\theta$ and decoder $D$ both have two hidden layers of size $512$ each. The parameter encoder $E_p$ maps a feature vector consisting of $p_i$ together with 40 sinusoidal encodings per parameter axis (sine and cosine at 20 log-spaced frequencies; 128 encodings for the H\'enon $u$-cascade) to the 16-dimensional latent space through a single hidden layer of size $64$. The predictor $P$ also contains a single hidden layer of width 64. Loss
hyperparameters are set to be $w_{\text{inv}} = w_{\text{var}} = 25$, $w_{\text{cov}} = 0$,
with $\gamma = 1$. Both stages use Adam at learning
rate $3\times10^{-4}$, batch size 256, and 5000 steps. We train 5 independent seeds per configuration.

\subsection{Analysis}
\label{sec:analysis}
We evaluate JEPAWG on four dynamical systems in two respects. First, we
decode weights at unseen parameters and compare the resulting forecasts
against the true trajectories and against the baselines. Accurate
forecasts show that the decoded weights retain the information specifying
the dynamics, which we then probe by measuring how strongly the decoded
weights respond to a small perturbation of the parameters and comparing
where this response peaks with the known bifurcations.
\paragraph{Forecasting}
We distinguish between \emph{interpolation} and \emph{extrapolation}, where we evaluate the forecaster at parameters $p$ inside or outside the training range, respectively. At a test parameter $p$ we decode
$\hat\theta(p) = D(P(E_p(p)))$ once and use the resulting network to
evolve a state for the recorded test length of that experiment (Table~\ref{tab:protocol}). Its accuracy can be measured by the normalised root-mean-square error
\begin{equation*}
\mathrm{NRMSE}(t) = \|\hat x(t) - x(t)\| \,/\, \sqrt{\mathrm{Var}[x]},
\end{equation*}
computed in the original state units (predictions are un-standardised
before evaluation). The normalising variance is that of the true test
trajectories at the same parameter, pooled over time, initial
conditions, and state dimensions:
$\mathrm{Var}[x] = \mathbb{E}_{x_0,t,d}\bigl[(x_d(t;x_0) -
\bar{x})^2\bigr]$ with $\bar{x}$ the pooled scalar mean. From this
metric we determine the \emph{time-to-threshold} (TtT): the longest initial duration over which the initial-condition-averaged NRMSE stays below
$0.2$. The TtT is averaged over the test parameters of
each task; all results are reported as mean $\pm$ standard deviation over the 5 training seeds.

\paragraph{Weight-space response}
\label{sec:proxy}
Beyond forecasting, JEPAWG lets us probe the dynamics through the learned
map $p \mapsto \hat\theta(p) = D(P(E_p(p)))$. We measure how fast the
decoded weights change with the control parameter,
\begin{equation}
    G(p) \;=\; \bigl\|\nabla_p\, \hat{\theta}(p)\bigr\|_F
    \;=\;
    \sqrt{\sum_{j=1}^{D_p}\sum_k
          \left(\partial \hat{\theta}_k / \partial p_j\right)^2},
    \label{eq:proxy}
\end{equation}
and expect it to peak at bifurcations. The reason is that a forecaster is
fitted where its training trajectories lie, approximately on the invariant
measure of the attractor. The vector fields studied here are smooth in $p$,
but the invariant measure reorganises when the attractor changes, and the
weights, which are pinned down on its support, change with it.
The derivative is evaluated by central differences on a dense grid of
evaluation points, which is independent of the parameters at which
networks were fine-tuned,
\begin{equation*}
    \partial \hat{\theta}_k / \partial p_j
    \;\simeq\;
    \bigl[\hat{\theta}_k(p + h e_j) - \hat{\theta}_k(p - h e_j)\bigr]/2h ,
\end{equation*}
with $e_j$ the unit vector along the $j$-th control parameter and $h$ the
evaluation step, and averaged over seeds, with per-experiment values of $h$ in Table~\ref{tab:protocol}. $G$ is computed in the standardised weight coordinates of Section~\ref{sec:weight-grid}; the raw-weight curve in Fig.~\ref{fig:lorenz-bif} (top) uses unstandardised weights. Note that by differentiating the \emph{decoded} weights, rather than the raw weights, we suppress per-parameter training noise and obtain a substantially cleaner signal (Fig.~\ref{fig:lorenz-bif}, top vs.\ middle plots).

What $G(p)$ is able to detect depends on the trajectories the forecasting networks
were trained on. Since the weights are pinned down only where these
trajectories lie, $G$ should peak at any transition that changes the
attracting dynamics. This includes local bifurcations such as the Hopf
and period-doubling events above, as well as global transitions that
alter the attractor, for which no eigenvalue condition exists
(Section~\ref{sec:setup}). Transitions that leave typical trajectories
unchanged, such as bifurcations of non-attracting sets, produce no
response. Two further properties are set by how the training dataset is
built. Its resolution is bounded from below by the average parameter
spacing (Table~\ref{tab:protocol}), so transitions that are closer to
each other than this effective width cannot be distinguished. And
whether transients are discarded or retained is a deliberate,
per-experiment choice (Table~\ref{tab:protocol}). Discarding them yields
weights that encode only on-attractor statistics (the standard
forecasting setting). Retaining them opens a different kind of analysis,
because the data then also sample the relaxation onto the attractor, which slows down critically
near a bifurcation, causing the weights to
carry a second signature beyond the change in the invariant measure.
 
\section{Experiments}
\label{sec:exp}
We evaluate on four parametric families, chosen so that each admits an independently known bifurcation structure as ground truth; parameter
ranges and grid sizes $M$ for all systems are listed in
Table~\ref{tab:all-inter-extrap} and supplementary details are given in Table~\ref{tab:protocol}. The \textbf{Lorenz} system,
\begin{equation}
    \dot{x} = \sigma(y - x),\qquad
    \dot{y} = x(\rho - z) - y,\qquad
    \dot{z} = xy - \beta z,
    \label{eq:lorenz}
\end{equation}
is a classical example of a system that can exhibit chaotic behaviour.
With $\sigma{=}10$, $\beta{=}8/3$ fixed, the system crosses a Hopf
bifurcation at $\rho_H\approx24.74$ beyond which the fixed points
lose stability and trajectories settle onto the strange attractor. When $\sigma$ is varied as well,
this point becomes the analytic Hopf curve
\begin{equation}\rho_H(\sigma)=\sigma(\sigma{+}\beta{+}3)/(\sigma{-}\beta{-}1).\label{eq:lorenzhopf} 
\end{equation}
    
We study both a one-parameter $\rho$-sweep and a two-parameter
$(\sigma,\rho)$-plane.

The \textbf{R\"ossler} oscillator,
\begin{equation}
    \dot{x} = -y - z,\qquad
    \dot{y} = x + a y,\qquad
    \dot{z} = b + z(x - c),
    \label{eq:rossler}
\end{equation}
is a minimal chaotic system with a single non-linear term. With
$a{=}b{=}0.2$ fixed, as $c$ increases the attractor undergoes
a period-doubling cascade, the doublings accumulating at the onset
of chaotic motion. 

The
\textbf{Brusselator},
\begin{equation}
    \dot{x} = a + x^2 y - (b{+}1)x,\qquad
    \dot{y} = bx - x^2 y,
    \label{eq:brusselator}
\end{equation}
is used as a model for autocatalytic reactions, and has a single fixed point at
$(a,\,b/a)$, which is stable only if $b<1+a^2$: a \textit{supercritical} Hopf curve. When crossed, the fixed point hands
over smoothly to a small stable limit cycle whose amplitude grows
continuously from zero, in contrast to the \textit{subcritical} Lorenz Hopf
above, where trajectories jump discontinuously onto the coexisting
strange attractor. We use a one-parameter $b$-sweep, with $a=1.25$ fixed, for forecasting. Finally, the discrete-time \textbf{H\'enon} map,
\begin{equation}
    x_{n+1} = 1 - a\,x_n^2 + y_n,\qquad
    y_{n+1} = b\,x_n,
    \label{eq:henon}
\end{equation}
is a standard two-dimensional chaotic map. As $a$ increases at fixed
$b$ it undergoes a period-doubling cascade, with successive doublings
at decreasing parameter intervals. Since the doubling locations are
known, this provides a direct test of the resolution of $G(p)$. We use
the map in three configurations, with different protocols
(Table~\ref{tab:protocol}): a one-parameter $a$-sweep for forecasting, a
two-parameter $(a,b)$ grid, and a separate zoo built along the cascade,
described in Section~\ref{sec:interp-henon}.

\subsection{Forecasting}
\label{sec:bench}

Table~\ref{tab:all-inter-extrap} reports the time-to-threshold
$\mathrm{TtT}_{0.2}$ (Section~\ref{sec:analysis}) for the four
systems. Additional results and experimental details are given in
Tables~\ref{tab:extrap_regimes},~\ref{tab:protocol} and \ref{tab:testsets}.

On interpolation, JEPAWG is tied with the PHLieNet hypernetwork on three systems and within two time units of it on the Brusselator, and outperforms the AE and VAE baselines on all four systems. The decoded
weights therefore retain the information specifying the dynamics at
parameters not seen in training, the precondition for the weight-space
analysis of Section~\ref{sec:interp}. We further observe that the decoded weights forecast better than the raw fine-tuned weights they were trained to reconstruct, in $\mathrm{TtT}$ on all four systems and most strongly on R\"ossler, though not always better than the shared base network (Table~\ref{tab:raw-vs-decoded}).

\begin{figure}[t]
    \centering
    \includegraphics[width=0.9\linewidth]{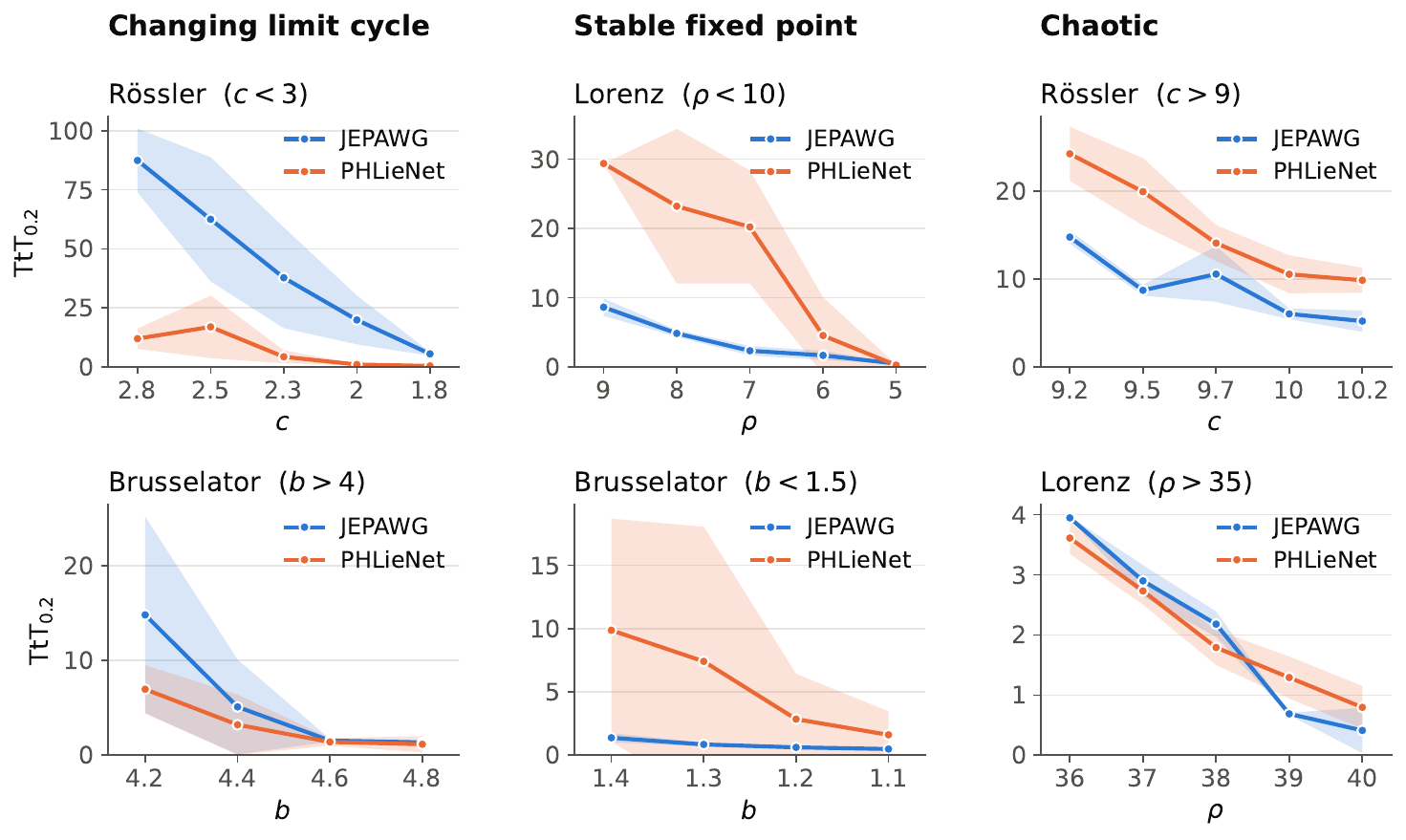}
\caption{Extrapolation $\mathrm{TtT}_{0.2}$ of JEPAWG and PHLieNet, by
regime beyond the training range. Left: limit cycles that keep changing;
for R\"ossler the cycle shrinks as $c$ decreases below the training range
($c<3$), for the Brusselator it grows and moves as $b$ increases above it
($b>4$). Middle: a stable fixed point, with initial transients included for
Lorenz and excluded for the Brusselator. Right: chaotic dynamics.}
    \label{fig:extrap_regimes}
\end{figure}

Extrapolation proves to be more challenging, and the performance is strongly dependent on the regime beyond the training range (Table~\ref{tab:extrap_regimes}). JEPAWG, AE and
VAE decode weights with MLPs that are unconstrained beyond the training
range and continue the trend of the weights past it. PHLieNet instead
saturates: as $p$ moves away from the anchors it returns the forecaster of the nearest
one. Where the exterior dynamics exhibit moving limit cycles, JEPAWG produces better forecasters: for R\"ossler with $c\in[1.8,2.8]$ the cycle keeps shrinking and JEPAWG holds a usable
forecast ($42.7\pm12.5$ against $7.0\pm2.3$), degrading monotonically with
distance, and the same ordering holds for the Brusselator with $b\in[4.2,4.8]$ until the
growing cycles leave the state-space region covered by training.
Where the dynamics beyond the range fall onto a fixed point, the forecaster
has to keep a zero of its vector field at a location that moves with the
parameter. For Lorenz with $\rho\in[5,9]$, the forecasters generated by PHLieNet preserve the fixed point more reliably than those decoded by JEPAWG ($15.5\pm4.2$ against $3.6\pm0.3$), whose vector field displaces or removes it. For the Brusselator with $b\in[1.1,1.4]$, trajectories after the initial transient lie
close to the fixed point, so the variance normalising the NRMSE is small and
even small offsets exceed the threshold. PHLieNet scores higher here, but with a near-zero normaliser and one seed dominating the mean, the difference is not meaningful.
Where the dynamics beyond the range are chaotic, small one-step errors grow
exponentially, so $\mathrm{TtT}$ is limited by the Lyapunov time rather than
by how well the attractor itself is reproduced. For R\"ossler with $c\in[9.2,10.2]$, PHLieNet leads, although its rollouts stay bounded slightly less often ($93\pm7\%$ against $100\%$;
Table~\ref{tab:extrap_regimes}). For Lorenz with $\rho\in[36,40]$, all methods reach
the threshold after about two to three Lyapunov times. Figure~\ref{fig:extrap_regimes} shows $\mathrm{TtT}_{0.2}$ for each regime as the parameter moves further outside the training range, and
Figure~\ref{fig:extrap_regimes_deepest} shows examples of predicted trajectories at the
furthest extrapolation parameter.

Accurate forecasts from the decoded networks show that the map
$p \mapsto \hat\theta(p)$ retains the information specifying the dynamics,
but this does not guarantee that this information is arranged in a way that can be retrieved by differentiation. PHLieNet
is such an example: it matches JEPAWG on forecasting
(Table~\ref{tab:all-inter-extrap}) yet its generated weights carry no visible response at the Lorenz Hopf point
(Section~\ref{sec:interp-lorenz}). JEPAWG instead fits the parameter--weight correspondence directly, which is
what the weight-space analysis relies on.

\begin{table}[t]
\centering
\caption{Forecasting performance: $\mathrm{TtT}_{0.2}$ on held-out
interpolation parameters and on each side of the training range, mean
$\pm$ std over 5 seeds. Bold: best mean and all entries whose interval overlaps the best entry’s. Rows in which all entries overlap carry no
bold. Continuous-time systems in time units, the discrete
H\'enon map in iterations; H\'enon extrapolation is omitted: for discrete
maps one inaccurate iteration exits the training region, so
$\mathrm{TtT}$ collapses to $\approx0$ for every method. Per-experiment
transient handling is given in Table~\ref{tab:protocol} and the full test
sets in Table~\ref{tab:testsets}.}
\label{tab:all-inter-extrap}
\footnotesize
\setlength{\tabcolsep}{4.5pt}
\renewcommand{\arraystretch}{1.05}
\begin{tabular}{llcccc}
\toprule
system & split & PHLieNet & JEPAWG & VAE & AE \\
\midrule
Lorenz, $M{=}800$
            & interp $\rho\in[10,35]$ & $\mathbf{20.23 \pm 0.89}$ & $\mathbf{20.19 \pm 0.15}$ & $18.69 \pm 0.27$ & $18.31 \pm 0.64$ \\
            & below $\rho\in[5,9]$    & $\mathbf{15.53 \pm 4.18}$ & $3.61 \pm 0.25$  & $5.58 \pm 0.81$  & $4.31 \pm 1.82$  \\
            & above $\rho\in[36,40]$  & $2.04 \pm 0.22$  & $2.03 \pm 0.09$  & $\mathbf{2.48 \pm 0.11}$  & $\mathbf{2.21 \pm 0.22}$  \\
\midrule
R\"ossler, $M{=}1000$
            & interp $c\in[3,9]$      & $\mathbf{50.16 \pm 1.93}$ & $\mathbf{47.42 \pm 0.90}$ & $35.84 \pm 2.58$ & $30.10 \pm 2.49$ \\
            & below $c\in[1.8,2.8]$   & $6.98 \pm 2.25$  & $\mathbf{42.66 \pm 12.53}$ & $\mathbf{27.64 \pm 12.20}$ & $10.44 \pm 8.30$ \\
            & above $c\in[9.2,10.2]$  & $\mathbf{15.72 \pm 1.71}$ & $9.06 \pm 0.92$  & $7.86 \pm 0.51$  & $6.09 \pm 0.54$  \\
\midrule
Brusselator, $M{=}1024$
            & interp $b\in[1.5,4]$    & $\mathbf{27.34 \pm 0.23}$ & $25.40 \pm 0.51$ & $20.47 \pm 0.46$ & $20.56 \pm 1.12$ \\
            & below $b\in[1.1,1.4]$ & $5.44 \pm 6.04$ & $0.84 \pm 0.18$  & $0.47 \pm 0.11$  & $0.89 \pm 0.25$  \\
            & above $b\in[4.2,4.8]$   & $3.19 \pm 1.08$  & $5.71 \pm 3.26$  & $3.26 \pm 0.82$  & $2.36 \pm 0.19$  \\
\midrule
H\'enon, $M{=}1024$
            & interp $a\in[0.85,1.10]$ & $\mathbf{650 \pm 126}$ & $\mathbf{772 \pm 85}$ & $594 \pm 90$ & $471 \pm 116$ \\
\bottomrule
\end{tabular}
\end{table}

\begin{table}[t]
\centering
\caption{Extrapolation by regime of the true dynamics beyond the training
range: $\mathrm{TtT}_{0.2}$ (short-horizon tracking) and bounded fraction (share of
rollouts that stay within $1.5\times$ the extent of the true test
trajectories over the full horizon), mean$\pm$std over 5 seeds of per-side means; bold as in Table~\ref{tab:all-inter-extrap}. Intervals as in Table~\ref{tab:testsets}; $\lambda_{\max}$ is the
largest Lyapunov exponent over the range, $0$ on limit cycles and the leading
Jacobian eigenvalue at fixed points. JEPAWG leads on both measures
where the limit cycle keeps changing; PHLieNet holds a fixed point that JEPAWG
loses on Lorenz; in chaos the measures split: PHLieNet tracks longer for R\"ossler with $c\in[9.2,10.2]$, but JEPAWG's rollouts always stay bounded.}
\label{tab:extrap_regimes}
\small
\setlength{\tabcolsep}{4pt}
\begin{tabular}{llccccc}
\toprule
 & & & \multicolumn{2}{c}{JEPAWG} & \multicolumn{2}{c}{PHLieNet} \\
\cmidrule(lr){4-5}\cmidrule(lr){6-7}
side & regime & $\lambda_{\max}$ & $\mathrm{TtT}_{0.2}$ & bounded (\%) & $\mathrm{TtT}_{0.2}$ & bounded (\%) \\
\midrule
R\"ossler, $c\in[1.8,2.8]$ & limit cycle & $0$ & $\mathbf{42.7 \pm 12.5}$ & $66 \pm 8$ & $7.0 \pm 2.3$ & $49 \pm 11$ \\
Brusselator, $b\in[4.2,4.8]$ & limit cycle & $0$ & $5.7 \pm 3.3$ & $\mathbf{100 \pm 0}$ & $3.2 \pm 1.1$ & $70 \pm 27$ \\
\midrule
Lorenz, $\rho\in[5,9]$ & stable fixed pt. & $-0.65$ & $3.6 \pm 0.3$ & $54 \pm 7$ & $\mathbf{15.5 \pm 4.2}$ & $\mathbf{97 \pm 6}$ \\
Brusselator, $b\in[1.1,1.4]$ & stable fixed pt. & $-0.58$ & $0.8 \pm 0.2$ & $0 \pm 0$ & $5.4 \pm 6.0$ & $0 \pm 0$ \\
\midrule
R\"ossler, $c\in[9.2,10.2]$ & chaotic & $0.11$ & $9.1 \pm 0.9$ & $100 \pm 0$ & $\mathbf{15.7 \pm 1.7}$ & $93 \pm 7$ \\
Lorenz, $\rho\in[36,40]$ & chaotic & $1.13$ & $2.03 \pm 0.09$ & $100 \pm 0$ & $2.04 \pm 0.22$ & $99 \pm 1$ \\
\bottomrule
\end{tabular}
\end{table}

\subsection{Bifurcation Detection}
\label{sec:interp}

We now apply the weight-space response $G(p)$ introduced in Section~\ref{sec:proxy} to
locate regime changes, comparing its peaks with bifurcations of known
location: Hopf points and curves in the Lorenz and Brusselator systems, the
period-doubling cascade of the H\'enon map, and the global bifurcation
structure around the Lorenz T-point. The R\"ossler cascade is treated in
Appendix~\ref{app:rossler}.

\subsubsection{Lorenz}
\label{sec:interp-lorenz}

\begin{figure}[t]
\centering
\begin{subfigure}[c]{0.46\textwidth}
    \includegraphics[width=\linewidth]{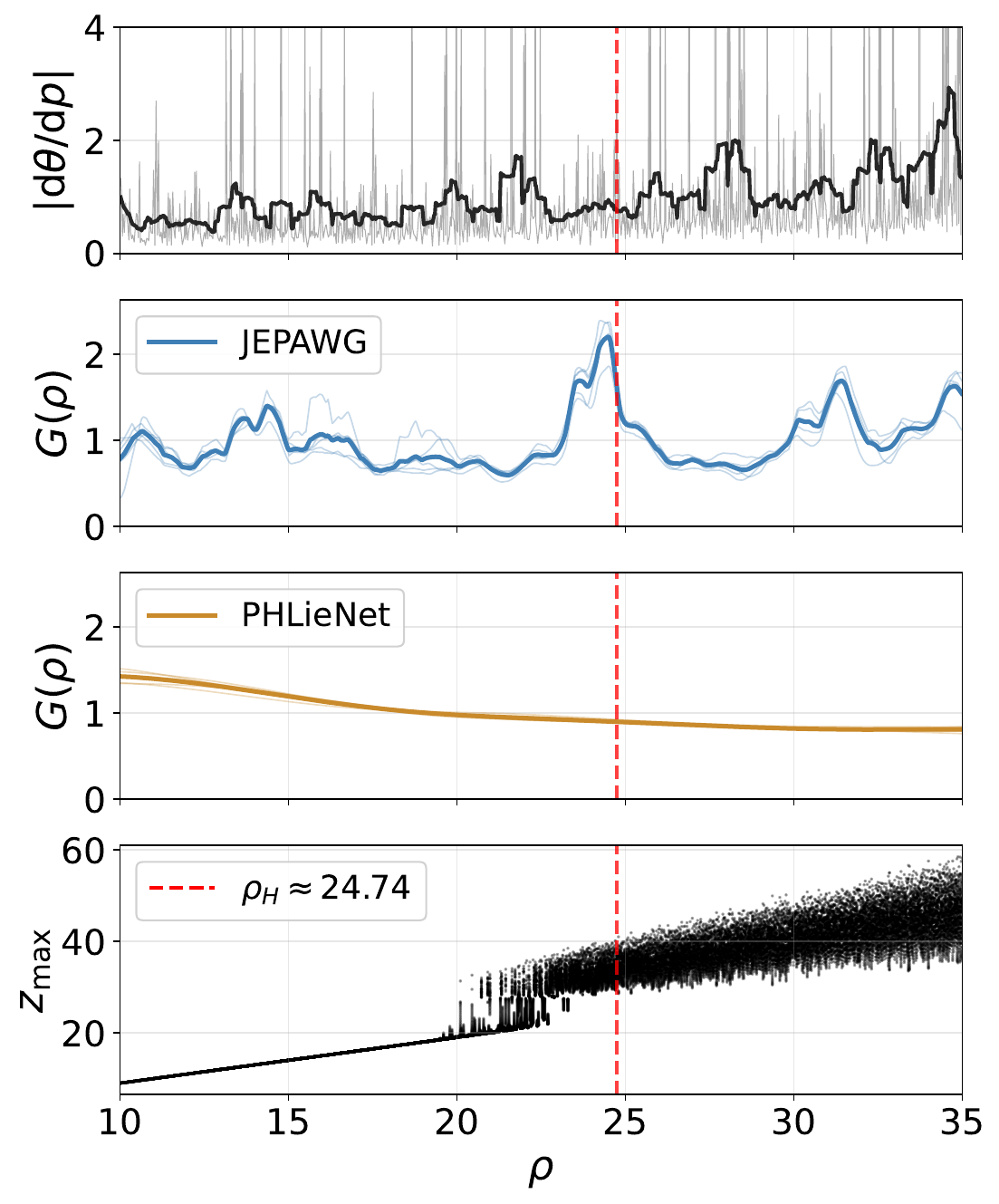}
    \caption{}
    \label{fig:lorenz-bif}
\end{subfigure}
\hfill
\begin{minipage}[c]{0.50\textwidth}
    \begin{subfigure}[t]{\linewidth}
        \centering
        \includegraphics[width=0.63\linewidth]{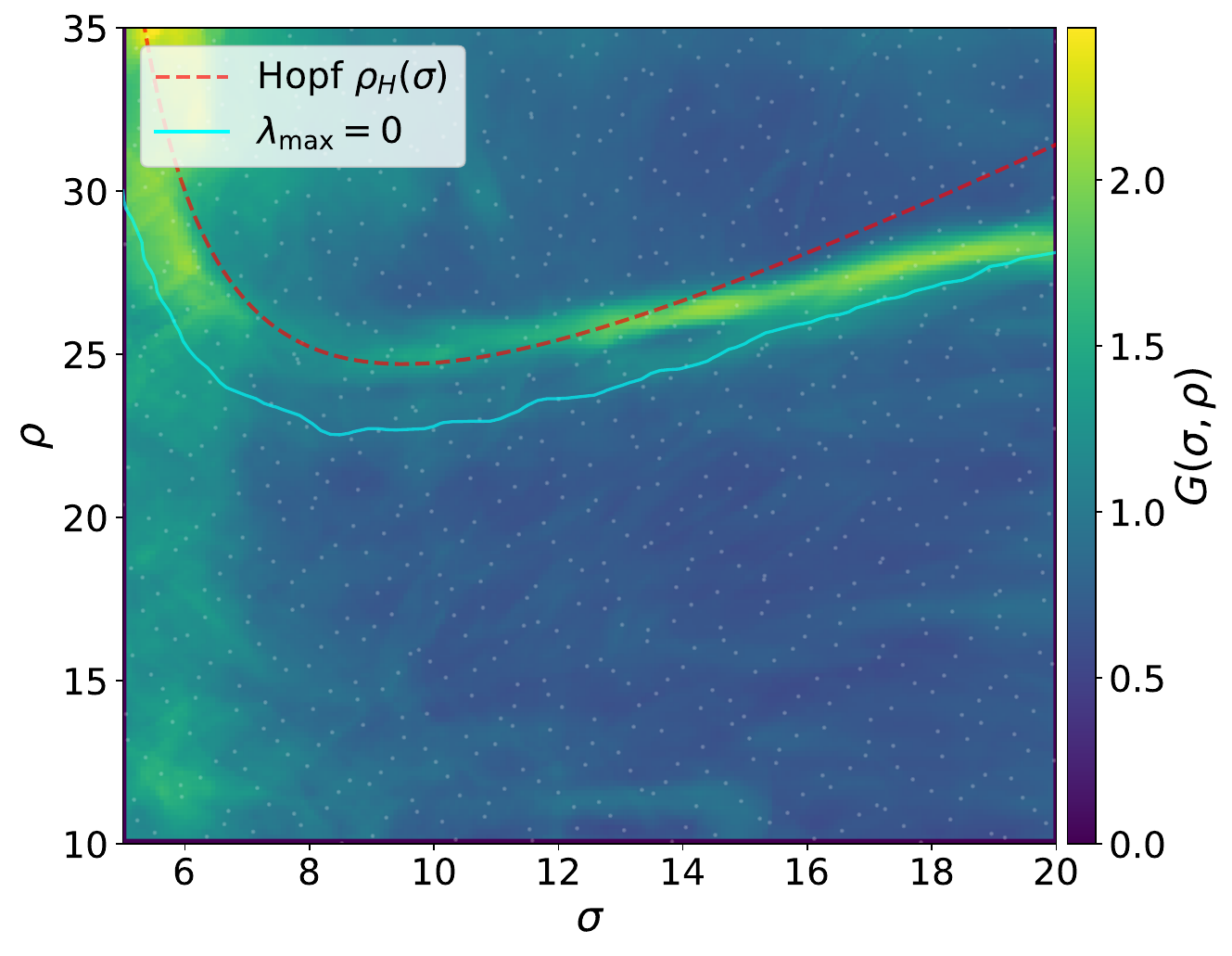}
        \caption{}
        \label{fig:lorenz2p-ridge}
    \end{subfigure}\\[4pt]
    \begin{subfigure}[t]{\linewidth}
        \centering
        \includegraphics[width=0.63\linewidth]{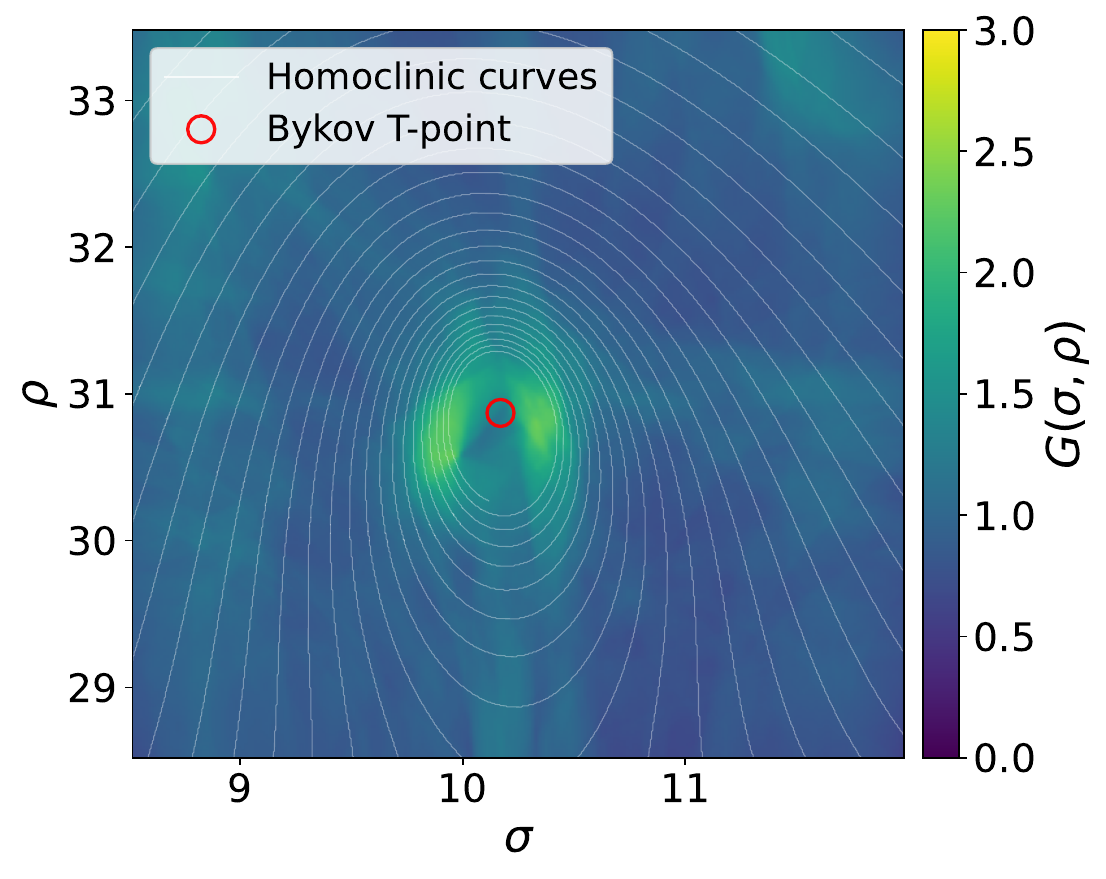}
        \caption{}
        \label{fig:lorenz-tpoint}
    \end{subfigure}
\end{minipage}
\caption{\textbf{Lorenz.} (a)~Raw-weights derivative $\|d\theta/d\rho\|$ across the $\rho$-sweep, with a
Gaussian-smoothed version as the thick line (top); derivative $G(\rho)=\|d\hat\theta/d\rho\|$ of the JEPAWG-decoded ($M{=}800$) and PHLieNet-generated weights, 5 seeds each, with their mean as the thick curves (middle); the ground-truth bifurcation diagram (bottom). The Hopf point $\rho_H$, at which the fixed points lose their stability, is shown as a red dashed line in all panels.
(b)~Weight-space response $G(\sigma,\rho)=\|\nabla_{\sigma,\rho}\hat\theta\|$
for a zoo trained on $(\sigma,\rho)\in[5,20]\times[10,40]$ ($M{=}1024$,
$\beta{=}8/3$), shown for $\rho\le35$ because the finite-difference
derivative is unreliable at the edge of the training range; the analytic
Hopf curve is in red, and the blue curve marks the level set
$\lambda_{\max}=0$ of the largest Lyapunov exponent.
(c)~Zoom above the Hopf curve with parameter ranges $(\sigma,\rho)\in[8.5,12]\times[28.5,33.5]$
($M{=}1024$, $\beta{=}8/3$): a ring of elevated response
surrounds the marked Bykov T-point at
$(\sigma,\rho)\approx(10.17,\,30.87)$. White lines: the principal
homoclinic curves of the origin, obtained independently of the model by
integrating its unstable manifold $W^u(0)$ on a $600^2$ grid and locating
the parameter values at which its winding number around the nearest equilibrium before
escaping to the other wing jumps by one turn. They form a single spiral
winding onto the T-point \citep{glendinning1986, Barrio2012KneadingsSD};
the ring of $G$ lies on its innermost turns, where successive curves are
closest, while no fixed point changes stability anywhere in the window.}
\label{fig:lorenz-interp}
\end{figure}

We first consider the Lorenz system, whose subcritical Hopf
bifurcation occurs at $\rho_H(\sigma)$ given in
\eqref{eq:lorenzhopf}. Varying only $\rho$ (at $\sigma=10$,
$\beta=8/3$), the bifurcation is a single point, and $G(\rho)$
shows a peak at $\rho = 24.51 \pm 0.23$ (mean and
standard deviation over five seeds), close
to the analytic $\rho_H \approx 24.74$. We do not expect the peak to coincide exactly with $\rho_H$: since the Hopf bifurcation is subcritical, the chaotic set already becomes attracting at $\rho \approx 24.06$ and coexists with the stable fixed points up to $\rho_H$, so the observed dynamics, and hence the trained weights, begin to change before the fixed points lose stability (Fig.~\ref{fig:lorenz-bif}, bottom). Neither the derivative of the raw weights nor that of the generated PHLieNet weights shows any discernible signal, in contrast to the JEPAWG-decoded weights (Figure~\ref{fig:lorenz-bif}). When also sweeping over $\sigma\in[5,20]$, the two-parameter response develops a ridge that tracks the Hopf curve for $\sigma < 13$ and moves towards the onset of chaos ($\lambda_{\max}=0$) for larger $\sigma$ (Figure~\ref{fig:lorenz2p-ridge}). Besides the Hopf curve and the onset of chaos, we probe the global bifurcation structure around the
Bykov T-point at $(\sigma,\rho)\approx(10.17,30.87)$ \cite{Pinsky_2017}, a
codimension-two heteroclinic point at which curves of homoclinic
bifurcations accumulate \citep{glendinning1986, Barrio2012KneadingsSD}.
Near it, small changes in $(\sigma,\rho)$ repeatedly rearrange the chaotic
dynamics, changing the pattern with which trajectories switch between the
two wings of the attractor, while no fixed point changes stability. As a result, the linearisation at the equilibria registers none of it.
Figure~\ref{fig:lorenz-tpoint} shows a ring of elevated response $G$
surrounding the T-point.

\subsubsection{Brusselator}
\label{sec:interp-brusselator}
\begin{figure}[t]
\centering
\begin{subfigure}[t]{0.48\textwidth}
    \includegraphics[width=0.8\linewidth]{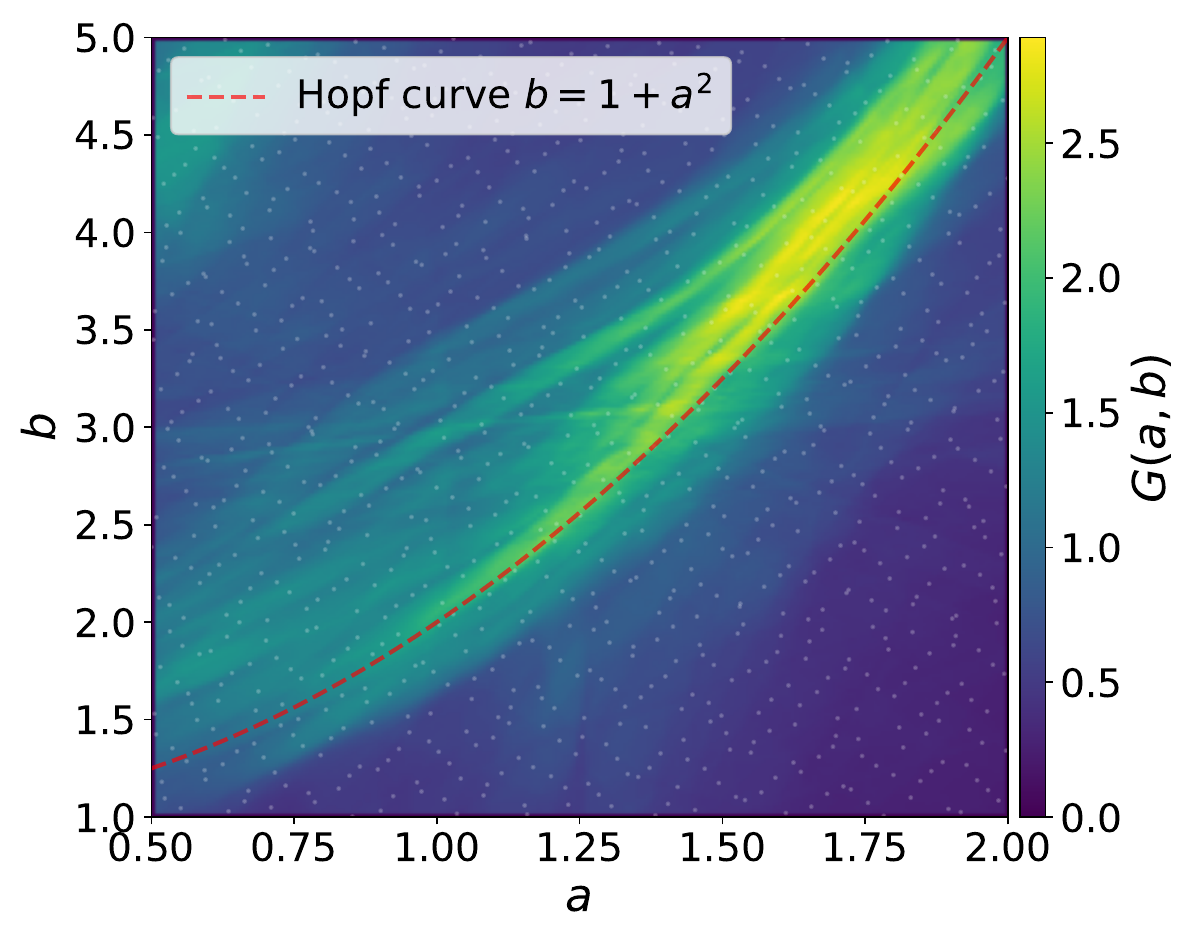}
    \caption{}
    \label{fig:brusselator-bif}
\end{subfigure}
\hfill
\begin{subfigure}[t]{0.48\textwidth}
    \includegraphics[width=0.8\linewidth]{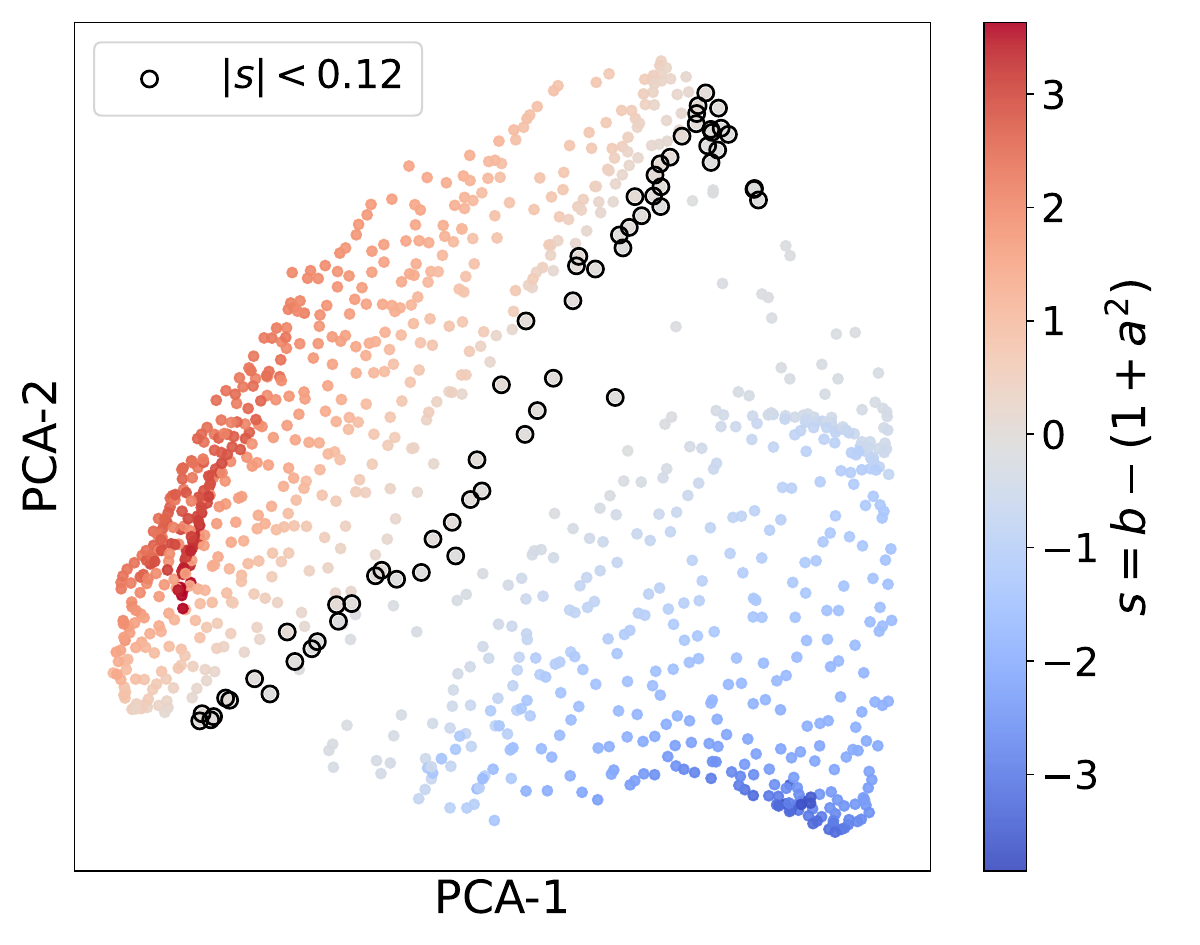}
    \caption{}
    \label{fig:brusselator-latent}
\end{subfigure}

\caption{\textbf{Brusselator.} (a)~Weight-space response $G(a,b) = \|\nabla_{a,b}\hat{\theta}\|$ with the analytic Hopf curve
$b=1+a^2$ in red; white dots mark the $M{=}1024$ Sobol sampled training parameters. (b)~Latent space (PCA-2D with $86\%$ explained variance), coloured by the signed distance
$s=b-(1+a^2)$ to the curve; circled points ($|s|<0.12$) correspond to the parameters nearest to the curve and trace a coherent boundary in every
seed. Parameter ranges are $(a,b)\in[0.5,2]\times[1,5]$ ($M{=}1024$).}
\label{fig:brusselator}
\end{figure}

The Brusselator displays a second two-parameter Hopf curve: whereas the Lorenz Hopf is
subcritical, with trajectories jumping onto the coexisting strange attractor,
crossing $b = 1 + a^2$ is supercritical, with the fixed point giving way to a
limit cycle that grows from zero amplitude.
With both $(a,b)$ varied, $G(p)$ again traces the analytic curve
(Fig.~\ref{fig:brusselator-bif}). The same transition is visible in
the learned latent space: the top two
principal components of the latent variable $z_w$ form a smoothly varying sheet in which
the two regimes occupy the two sides of an internal boundary, and
the sheet is stretched near the Hopf curve
(Fig.~\ref{fig:brusselator-latent}).

\subsubsection{H\'enon}
\label{sec:interp-henon}
Last, we consider the discrete-time H\'enon map \eqref{eq:henon},
which exhibits a period-doubling cascade as $a$ increases at fixed
$b=0.3$: the attractor's period doubles at a sequence of parameter
values $a_1 < a_2 < \dots$ accumulating at the chaos onset
$a_\infty \approx 1.05805$, visible as successive branchings in the
bifurcation diagram of Figure~\ref{fig:henon-perioddoubling}. The
doublings occur at shrinking intervals whose ratio converges to the
universal Feigenbaum constant \cite{collet1981, feigenbaum1978}
\begin{equation}
    \delta \;=\; \lim_{n\to\infty}\frac{a_n - a_{n-1}}{a_{n+1} - a_n}
    \;\approx\; 4.6692.
    \label{eq:feigenbaum}
\end{equation} 
In the following we test whether we can detect the doubling
locations with the weight-space response $G$, and qualitatively
recover the Feigenbaum scaling. The asymptotic progression of the doubling locations towards $a_\infty$
suggests a natural coordinate,
\begin{equation}
    u \;=\; \log_{10}\!\bigl(a_\infty - a\bigr),
    \label{eq:ucoord}
\end{equation}
in which the doublings are asymptotically equidistant, so that a sample
evenly spread in $u$ resolves every level with the same relative
precision. Concretely, we build a separate zoo for the cascade:
$M{=}1540$ parameters at $b{=}0.3$ over
$a\in[0.30,\,a_\infty{-}3{\times}10^{-5}]$, stratified in $u$, with $u$ as the control input and the initial transient retained
(Table~\ref{tab:protocol}).

\begin{figure}[t]
\centering
\begin{minipage}[c]{0.58\textwidth}
    \centering
    \includegraphics[width=\linewidth]{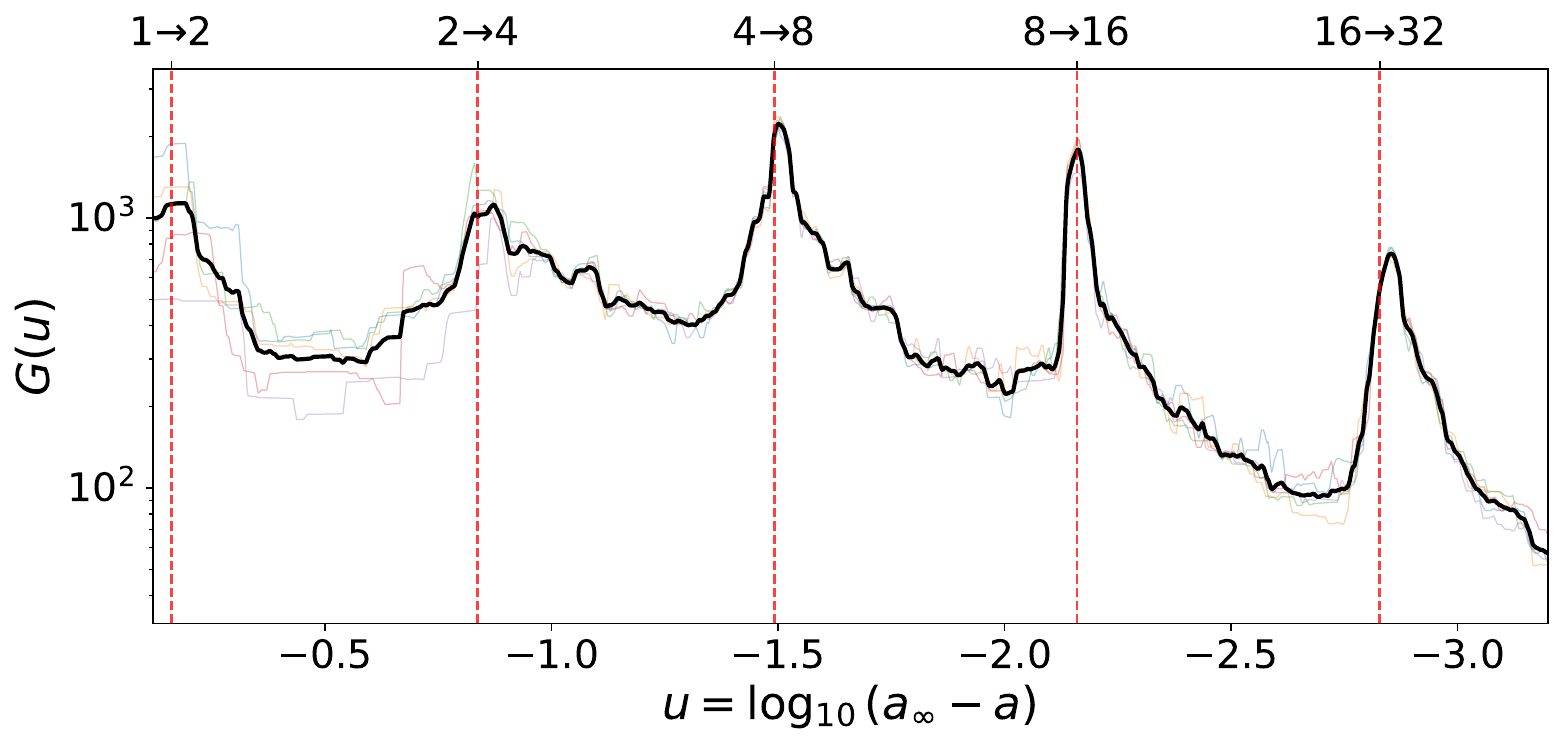}
    \captionof{figure}{Weight-space response $G(u)$ along the
    H\'enon $u$-cascade for $u>-3.2$, 5 seeds in colour and their mean in black. In order to include the dynamics of falling into the attractor, all transients are kept during training. The response and bifurcation diagram over the full sampled range are shown in Fig.~\ref{fig:henon-perioddoubling}
(Appendix~\ref{app:henon}).}
    \label{fig:henon-ucascade}
\end{minipage}
\hfill
\begin{minipage}[c]{0.35\textwidth}
    \centering
    \captionof{table}{Detected doubling locations for the H\'enon map (5-seed mean $G(u)$, local peaks)
    against bisection references.}
    \label{tab:henon-doublings}
    \footnotesize
    \begin{tabular}{lcc}
    \toprule
    event & reference $a$ & detected $a$ \\
    \midrule
    $1\to 2$    & $0.36749$ & $0.4159$  \\
    $2\to 4$    & $0.91250$ & $0.9242$  \\
    $4\to 8$    & $1.02585$ & $1.02649$ \\
    $8\to 16$   & $1.05112$ & $1.05116$ \\
    $16\to 32$  & $1.05656$ & $1.05665$ \\
    \bottomrule
    \end{tabular}
\end{minipage}
\end{figure}
$G(u)$ exhibits a local peak at each of the first five period doublings (Fig.~\ref{fig:henon-ucascade},
Table~\ref{tab:henon-doublings}), each localised to
within $|\Delta u| \le 0.04$. The estimated ratios
$(\hat\delta_1,\hat\delta_2,\hat\delta_3) = (4.97,\,4.15,\,4.49)$
are comparable to the reference ratios $(4.81,\,4.49,\,4.65)$, which themselves converge slowly to $\delta$. A universal constant of the underlying dynamics is thus qualitatively measured from the parameter derivative of the decoded weights.
On the two-parameter $(a,b)$ grid the ridge of $G$ follows the $2\!\to\!4$ and $4\!\to\!8$ doubling curves of the ground-truth period map and broadens over the chaotic band (Appendix~\ref{app:henon}, Fig.~\ref{fig:henon-period}).

\section{Conclusion}
For four dynamical systems, we built model zoos of fine-tuned forecasting
networks indexed by the control parameters, and evaluated the
parameter-to-weights map JEPAWG learns from them both as a hypernetwork and
as a probe of the dynamics. As a hypernetwork,
JEPAWG is competitive with an end-to-end baseline on
held-out interpolation parameters across all four systems.
Outside the training range, which method performs better depends on the
dynamics there. Where the limit cycles continue to change, JEPAWG follows
the trend of the weights and gives the better forecasts, most clearly for R\"ossler with $c\in[1.8,2.8]$. Where the dynamics settle onto a fixed point, PHLieNet,
which returns the forecaster of the nearest training parameter, preserves it
more reliably. In chaotic regimes, both methods lose the forecast within a few Lyapunov times. 
As a probe, the parameter derivative of the
decoded weights, $G(p)$, recovers the subcritical Hopf point and
curve of the Lorenz system and the supercritical Hopf curve of the
Brusselator, and registers the global bifurcation
structure around the Bykov T-point, which, unlike the Hopf curves, is invisible to any linearisation. Finally, we used the weight-space response $G(p)$ to localise the first five period
doublings of the H\'enon map to within $|\Delta u| \le 0.04$,
yielding a qualitative measurement of the Feigenbaum constant.
JEPAWG thus turns a parametrised model zoo into a single object
serving two roles: a hypernetwork producing functional forecasters
at unseen parameters, and a probe of the dynamics that generated
the zoo. Having validated the weight-space response against known bifurcation structure, we see a promising prospect in applying the method
to systems whose dynamics are not yet fully understood, where the
weights would point to structures not yet identified,
indicating where to concentrate simulation or further training
rather than sweeping a parameter range at uniform cost.

\paragraph{Limitations}
First, the probe can only register what is present in the training data.
Transitions that leave typical trajectories unchanged, such as bifurcations
of non-attracting sets, are invisible to it. Second, outside the range of
the training parameters nothing constrains the decoded weights, and how quickly the error destroys the forecast depends on the dynamics there. In
chaotic regimes it is amplified, and every method we tested loses the
forecast within a few Lyapunov times. Where the dynamics settle onto a
fixed point, JEPAWG continues the trend of the weights and displaces or
removes the fixed point (Section~\ref{sec:bench}). Third, $G(p)$ responds to any rapid change in the distribution of the
training trajectories, so a peak indicates a reorganisation of the observed
dynamics, not necessarily a bifurcation. Its resolution is bounded by the
average parameter spacing (Table~\ref{tab:protocol}) and by the smoothness
of the decoded map, and since $G$ is a derivative, the location of its
peaks depends on the coordinates in which the parameter is expressed.
Finally, the dominant cost is the zoo itself, which requires $M$
fine-tuned networks per system.

\section*{Acknowledgments}
T.~G. was supported by the AI-application grant AS-IAIA-114-M02 from Academia Sinica. M.~C.~N.~C. was supported by the same grant and by the Vici grant VI.C.232.117 from the Dutch Research Council (NWO).

\bibliographystyle{unsrtnat}
\bibliography{refs}

@article{vlachas2026phlienet,
  title = {Beyond Static Models: {{Hypernetworks}} for Adaptive and Generalizable Forecasting in Complex Parametric Dynamical Systems},
  author = {Vlachas, Pantelis R. and Vlachas, Konstantinos and Chatzi, Eleni},
  year = 2026,
  journal = {Mechanical Systems and Signal Processing},
  volume = {258},
  pages = {114635},
  issn = {0888-3270},
  doi = {10.1016/j.ymssp.2026.114635}
}

@InProceedings{navon2023equivariant,
  title = 	 {Equivariant Architectures for Learning in Deep Weight Spaces},
  author =       {Navon, Aviv and Shamsian, Aviv and Achituve, Idan and Fetaya, Ethan and Chechik, Gal and Maron, Haggai},
  booktitle = 	 {Proceedings of the 40th International Conference on Machine Learning},
  pages = 	 {25790--25816},
  year = 	 {2023},
  editor = 	 {Krause, Andreas and Brunskill, Emma and Cho, Kyunghyun and Engelhardt, Barbara and Sabato, Sivan and Scarlett, Jonathan},
  volume = 	 {202},
  series = 	 {Proceedings of Machine Learning Research},
  month = 	 {23--29 Jul},
  publisher =    {PMLR},
  url = 	 {https://proceedings.mlr.press/v202/navon23a.html}
}

@inproceedings{zhou2023neuralfunctionals,
  title = {Permutation Equivariant Neural Functionals},
  booktitle = {Advances in Neural Information Processing Systems},
  author = {Zhou, Allan and Yang, Kaien and Burns, Kaylee and Cardace, Adriano and Jiang, Yiding and Sokota, Samuel and Kolter, J. Zico and Finn, Chelsea},
  editor = {Oh, A. and Naumann, T. and Globerson, A. and Saenko, K. and Hardt, M. and Levine, S.},
  year = {2023},
  volume = {36},
  pages = {24966--24992},
  publisher = {Curran Associates, Inc.},
  doi = {10.52202/075280-1085},
  url = {https://proceedings.neurips.cc/paper_files/paper/2023/file/4e9d8aeeab6120c3c83ccf95d4c211d3-Paper-Conference.pdf}
}

@InProceedings{assran2023ijepa,
    author    = {Assran, Mahmoud and Duval, Quentin and Misra, Ishan and Bojanowski, Piotr and Vincent, Pascal and Rabbat, Michael and LeCun, Yann and Ballas, Nicolas},
    title     = {Self-Supervised Learning From Images With a Joint-Embedding Predictive Architecture},
    booktitle = {Proceedings of the IEEE/CVF Conference on Computer Vision and Pattern Recognition (CVPR)},
    month     = {June},
    year      = {2023},
    pages     = {15619-15629}
}

@misc{bardes2022vicreg,
      title={VICReg: Variance-Invariance-Covariance Regularization for Self-Supervised Learning}, 
      author={Adrien Bardes and Jean Ponce and Yann LeCun},
      year={2022},
      eprint={2105.04906},
      archivePrefix={arXiv},
      primaryClass={cs.CV},
      url={https://arxiv.org/abs/2105.04906}, 
}

@article{vannieuwenburg2017learning,
  title = {Learning Phase Transitions by Confusion},
  author = {{van Nieuwenburg}, Evert~P.~L. and Liu, Ye-Hua and Huber, Sebastian~D.},
  year = 2017,
  month = may,
  journal = {Nature Physics},
  volume = {13},
  number = {5},
  pages = {435--439},
  issn = {1745-2481},
  doi = {10.1038/nphys4037}
}

@misc{jepawg,
      title={Weight-Space Physics: Interpretable Hypernetworks for Lattice Quantum Field Theories}, 
      author={Tobias Göbel and Julian R. Ebelt and Zier Mensch and Mathis Gerdes and Miranda C. N. Cheng},
      year={2026},
      eprint={2607.07127},
      archivePrefix={arXiv},
      primaryClass={hep-lat},
      url={https://arxiv.org/abs/2607.07127}, 
}

@misc{graves2014generatingsequencesrecurrentneural,
      title={Generating Sequences With Recurrent Neural Networks}, 
      author={Alex Graves},
      year={2014},
      eprint={1308.0850},
      archivePrefix={arXiv},
      primaryClass={cs.NE},
      url={https://arxiv.org/abs/1308.0850}, 
}

@misc{vlachas2020backpropagationalgorithmsreservoircomputing,
      title={Backpropagation Algorithms and Reservoir Computing in Recurrent Neural Networks for the Forecasting of Complex Spatiotemporal Dynamics}, 
      author={Pantelis R. Vlachas and Jaideep Pathak and Brian R. Hunt and Themistoklis P. Sapsis and Michelle Girvan and Edward Ott and Petros Koumoutsakos},
      year={2020},
      eprint={1910.05266},
      archivePrefix={arXiv},
      primaryClass={eess.SP},
      url={https://arxiv.org/abs/1910.05266}, 
}

@misc{duth2026mechanisticanalysistransformersdynamical,
      title={A Mechanistic Analysis of Transformers for Dynamical Systems}, 
      author={Gregory Duthé and Nikolaos Evangelou and Wei Liu and Ioannis G. Kevrekidis and Eleni Chatzi},
      year={2026},
      eprint={2512.21113},
      archivePrefix={arXiv},
      primaryClass={cs.LG},
      url={https://arxiv.org/abs/2512.21113}, 
}

@misc{bai2018empiricalevaluationgenericconvolutional,
      title={An Empirical Evaluation of Generic Convolutional and Recurrent Networks for Sequence Modeling}, 
      author={Shaojie Bai and J. Zico Kolter and Vladlen Koltun},
      year={2018},
      eprint={1803.01271},
      archivePrefix={arXiv},
      primaryClass={cs.LG},
      url={https://arxiv.org/abs/1803.01271}, 
}

@article{Pinsky_2017,
  title = {On the Topology of the {{Lorenz}} System},
  author = {Pinsky, Tali},
  year = 2017,
  month = sep,
  journal = {Proceedings of the Royal Society A: Mathematical, Physical and Engineering Sciences},
  volume = {473},
  number = {2205},
  eprint = {https://royalsocietypublishing.org/rspa/article-pdf/doi/10.1098/rspa.2017.0374/365330/rspa.2017.0374.pdf},
  pages = {20170374},
  issn = {1364-5021},
  doi = {10.1098/rspa.2017.0374}
}

@article{feigenbaum1978,
  title = {Quantitative Universality for a Class of Nonlinear Transformations},
  author = {Feigenbaum, Mitchell J.},
  year = 1978,
  month = jul,
  journal = {Journal of Statistical Physics},
  volume = {19},
  number = {1},
  pages = {25--52},
  issn = {1572-9613},
  doi = {10.1007/BF01020332}
}

@article{collet1981,
  title = {Period Doubling Bifurcations for Families of Maps on {$\mathbb{R}^n$}},
  author = {Collet, P. and Eckmann, J. -P. and Koch, H.},
  year = 1981,
  month = may,
  journal = {Journal of Statistical Physics},
  shortjournal = {Journal of Statistical Physics},
  volume = {25},
  number = {1},
  pages = {1--14},
  issn = {1572-9613},
  doi = {10.1007/BF01008475},
  url = {https://doi.org/10.1007/BF01008475}
}

@article{glendinning1986,
  title = {T-Points: {{A}} Codimension Two Heteroclinic Bifurcation},
  author = {Glendinning, Paul and Sparrow, Colin},
  year = 1986,
  month = may,
  journal = {Journal of Statistical Physics},
  volume = {43},
  number = {3},
  pages = {479--488},
  issn = {1572-9613},
  doi = {10.1007/BF01020649}
}

@article{Barrio2012KneadingsSD,
  title   = {Kneadings, Symbolic Dynamics and Painting {Lorenz} Chaos},
  author  = {Barrio, Roberto and Shilnikov, Andrey and Shilnikov, Leonid},
  journal = {International Journal of Bifurcation and Chaos},
  volume  = {22},
  number  = {04},
  pages   = {1230016},
  year    = {2012},
  doi     = {10.1142/S0218127412300169}
}

@book{strogatz:2000,
  title     = {Nonlinear Dynamics and Chaos: With Applications to Physics, Biology, Chemistry and Engineering},
  author    = {Strogatz, Steven H.},
  publisher = {Westview Press},
  year      = {2000}
}

@article{facco2017estimating,
  title = {Estimating the Intrinsic Dimension of Datasets by a Minimal Neighborhood Information},
  author = {Facco, Elena and {d'Errico}, Maria and Rodriguez, Alex and Laio, Alessandro},
  year = 2017,
  month = sep,
  journal = {Scientific Reports},
  volume = {7},
  number = {1},
  pages = {12140},
  issn = {2045-2322},
  doi = {10.1038/s41598-017-11873-y}
}

@inproceedings{levina2004maximum,
  title = {Maximum Likelihood Estimation of Intrinsic Dimension},
  booktitle = {Advances in Neural Information Processing Systems},
  author = {Levina, Elizaveta and Bickel, Peter},
  editor = {Saul, L. and Weiss, Y. and Bottou, L.},
  year = 2004,
  volume = {17},
  publisher = {MIT Press}
}

@misc{wortsman2022modelsoupsaveragingweights,
      title={Model soups: averaging weights of multiple fine-tuned models improves accuracy without increasing inference time}, 
      author={Mitchell Wortsman and Gabriel Ilharco and Samir Yitzhak Gadre and Rebecca Roelofs and Raphael Gontijo-Lopes and Ari S. Morcos and Hongseok Namkoong and Ali Farhadi and Yair Carmon and Simon Kornblith and Ludwig Schmidt},
      year={2022},
      eprint={2203.05482},
      archivePrefix={arXiv},
      primaryClass={cs.LG},
      url={https://arxiv.org/abs/2203.05482}, 
}

@misc{schürholt2022modelzoosdatasetdiverse,
      title={Model Zoos: A Dataset of Diverse Populations of Neural Network Models}, 
      author={Konstantin Schürholt and Diyar Taskiran and Boris Knyazev and Xavier Giró-i-Nieto and Damian Borth},
      year={2022},
      eprint={2209.14764},
      archivePrefix={arXiv},
      primaryClass={cs.LG},
      url={https://arxiv.org/abs/2209.14764}, 
}

@book{guckenheimer1983nonlinear,
  title={Nonlinear Oscillations, Dynamical Systems, and Bifurcations of Vector Fields},
  author={Guckenheimer, John and Holmes, Philip},
  volume={42},
  series={Applied Mathematical Sciences},
  year={1983},
  publisher={Springer New York},
  doi={10.1007/978-1-4612-1140-2}
}

@article{6795963,
  title = {Long Short-Term Memory},
  author = {Hochreiter, Sepp and Schmidhuber, J{\"u}rgen},
  year = 1997,
  month = nov,
  journal = {Neural Computation},
  volume = {9},
  number = {8},
  eprint = {https://direct.mit.edu/neco/article-pdf/9/8/1735/813796/neco.1997.9.8.1735.pdf},
  pages = {1735--1780},
  issn = {0899-7667},
  doi = {10.1162/neco.1997.9.8.1735}
}

@article{Sobol1967OnTD,
  title = {On the Distribution of Points in a Cube and the Approximate Evaluation of Integrals},
  author = {Sobol', I.M},
  year = 1967,
  journal = {USSR Computational Mathematics and Mathematical Physics},
  volume = {7},
  number = {4},
  pages = {86--112},
  issn = {0041-5553},
  doi = {10.1016/0041-5553(67)90144-9}
}

@article{tidhy,
  title = {{{TiDHy}}: Timescale Demixing via Hypernetworks to Learn Simultaneous Dynamics from Mixed Observations},
  author = {Abe, Elliott Taylor Tsuyoshi and Brunton, Bingni W.},
  year = 2026,
  month = may,
  journal = {Journal of The Royal Society Interface},
  volume = {23},
  number = {238},
  pages = {20250738},
  issn = {1742-5689},
  doi = {10.1098/rsif.2025.0738},
  urldate = {2026-10-18}
}

@misc{nqsweights,
      title={Adiabatic Fine-Tuning of Neural Quantum States Enables Detection of Phase Transitions in Weight Space}, 
      author={Vinicius Hernandes and Thomas Spriggs and Saqar Khaleefah and Eliska Greplova},
      year={2025},
      eprint={2503.17140},
      archivePrefix={arXiv},
      primaryClass={quant-ph},
      url={https://arxiv.org/abs/2503.17140}, 
}

@article{mogharabin2025bifurcation,
  title = {Bifurcation Analysis in Dynamical Systems through Integration of Machine Learning and Dynamical Systems Theory},
  author = {Mogharabin, Nami and Ghadami, Amin},
  year = 2024,
  month = dec,
  journal = {Journal of Computational and Nonlinear Dynamics},
  volume = {20},
  number = {2},
  eprint = {https://asmedigitalcollection.asme.org/computationalnonlinear/article-pdf/20/2/021006/7420307/cnd\_020\_02\_021006.pdf},
  pages = {021006},
  issn = {1555-1415},
  doi = {10.1115/1.4067297}
}

@misc{pal2026machinelearningcomplexsystems,
      title={Machine Learning for Complex Systems Dynamics: Detecting Bifurcations in Dynamical Systems with Deep Neural Networks}, 
      author={Swadesh Pal and Roderick Melnik},
      year={2026},
      eprint={2603.04420},
      archivePrefix={arXiv},
      primaryClass={cs.LG},
      url={https://arxiv.org/abs/2603.04420}, 
}

@misc{modelhubs,
      title={Learning Model Representations Using Publicly Available Model Hubs}, 
      author={Damian Falk and Konstantin Schürholt and Konstantinos Tzevelekakis and Léo Meynent and Damian Borth},
      year={2025},
      eprint={2510.02096},
      archivePrefix={arXiv},
      primaryClass={cs.LG},
      url={https://arxiv.org/abs/2510.02096}, 
}

@misc{ha2016hypernetworks,
      title={HyperNetworks}, 
      author={David Ha and Andrew Dai and Quoc V. Le},
      year={2016},
      eprint={1609.09106},
      archivePrefix={arXiv},
      primaryClass={cs.LG},
      url={https://arxiv.org/abs/1609.09106}, 
}

@InProceedings{kirchmeyer2022generalizingnewphysicalsystems,
  title = 	 {Generalizing to New Physical Systems via Context-Informed Dynamics Model},
  author =       {Kirchmeyer, Matthieu and Yin, Yuan and Dona, Jeremie and Baskiotis, Nicolas and Rakotomamonjy, Alain and Gallinari, Patrick},
  booktitle = 	 {Proceedings of the 39th International Conference on Machine Learning},
  pages = 	 {11283--11301},
  year = 	 {2022},
  editor = 	 {Chaudhuri, Kamalika and Jegelka, Stefanie and Song, Le and Szepesvari, Csaba and Niu, Gang and Sabato, Sivan},
  volume = 	 {162},
  series = 	 {Proceedings of Machine Learning Research},
  month = 	 {17--23 Jul},
  publisher =    {PMLR},
  url = 	 {https://proceedings.mlr.press/v162/kirchmeyer22a.html}
}

@article{Tanaka_2017,
   title={Detection of Phase Transition via Convolutional Neural Networks},
   volume={86},
   ISSN={1347-4073},
   url={http://dx.doi.org/10.7566/JPSJ.86.063001},
   DOI={10.7566/jpsj.86.063001},
   number={6},
   journal={Journal of the Physical Society of Japan},
   publisher={Physical Society of Japan},
   author={Tanaka, Akinori and Tomiya, Akio},
   year={2017},
   month=jun, pages={063001} }

@misc{Ranger,
  author = {Wright, Less},
  title = {Ranger - a synergistic optimizer.},
  year = {2019},
  publisher = {GitHub},
  journal = {GitHub repository},
  howpublished = {\url{https://github.com/lessw2020/Ranger-Deep-Learning-Optimizer}}
}

@inproceedings{NIPS2017_3f5ee243,
  title = {Attention Is All You Need},
  booktitle = {Advances in Neural Information Processing Systems},
  author = {Vaswani, Ashish and Shazeer, Noam and Parmar, Niki and Uszkoreit, Jakob and Jones, Llion and Gomez, Aidan N and Kaiser, {\L}ukasz and Polosukhin, Illia},
  editor = {Guyon, I. and Luxburg, U. Von and Bengio, S. and Wallach, H. and Fergus, R. and Vishwanathan, S. and Garnett, R.},
  year = 2017,
  volume = {30},
  publisher = {Curran Associates, Inc.}
}

@article{Carrasquilla_2017,
   title={Machine learning phases of matter},
   volume={13},
   ISSN={1745-2481},
   url={http://dx.doi.org/10.1038/nphys4035},
   DOI={10.1038/nphys4035},
   number={5},
   journal={Nature Physics},
   publisher={Springer Science and Business Media LLC},
   author={Carrasquilla, Juan and Melko, Roger G.},
   year={2017},
   month=Feb, pages={431–434} }

@article{Bury_2023,
   title={Predicting discrete-time bifurcations with deep learning},
   volume={14},
   ISSN={2041-1723},
   url={http://dx.doi.org/10.1038/s41467-023-42020-z},
   DOI={10.1038/s41467-023-42020-z},
   number={1},
   journal={Nature Communications},
   publisher={Springer Science and Business Media LLC},
   author={Bury, Thomas M. and Dylewsky, Daniel and Bauch, Chris T. and Anand, Madhur and Glass, Leon and Shrier, Alvin and Bub, Gil},
   year={2023},
   month=Oct }

@article{bury2021deep,
  title = {Deep Learning for Early Warning Signals of Tipping Points},
  author = {Bury, Thomas M. and Sujith, R. I. and Pavithran, Induja and Scheffer, Marten and Lenton, Timothy M. and Anand, Madhur and Bauch, Chris T.},
  year = 2021,
  journal = {Proceedings of the National Academy of Sciences},
  volume = {118},
  number = {39},
  eprint = {https://www.pnas.org/doi/pdf/10.1073/pnas.2106140118},
  pages = {e2106140118},
  doi = {10.1073/pnas.2106140118}
}

\appendix
\section{Architectural Details}
\paragraph{Forecasting network}
The forecasting network throughout is the TCNN-CD~\cite{bai2018empiricalevaluationgenericconvolutional} in the configuration of \cite{vlachas2026phlienet}: a causal temporal CNN with dilations
$2^0,\dots,2^{L-1}$ ($L$ chosen as the minimum depth whose receptive
field covers the input window, $L{=}4$ at input sequence length $T=32$),
$22$ channels per layer, kernel size $5$, and a linear readout. PHLieNet uses the same TCNN-CD as its generated target
network, with a LIE of $32$ anchors, embedding dimension $16$, RBF bandwidth $\ell{=}0.2$, and a hypernetwork MLP of two hidden layers
of width $64$; it is trained with Ranger (RAdam +
Lookahead) \cite{Ranger}, initial learning rate $10^{-2}$, and the training protocol
of \cite{vlachas2026phlienet}.

\paragraph{Baselines}
We compare JEPAWG's embedding against PCA, a deterministic
autoencoder (AE), and a variational autoencoder (VAE) on the weight grids of all four systems (layer sizes below are for Lorenz, $M{=}800$). All
methods use latent dimension $d_z{=}16$ and weights normalised
per-feature with training-split statistics. PCA projects onto the
top 16 training principal components; AE/VAE mirror JEPAWG's
encoder/decoder capacity ($7857{\to}512{\to}512{\to}16$ and back;
VAE with $\beta_{\mathrm{KL}}{=}10^{-3}$). Since PCA/AE/VAE provide no
parameter-to-latent map, each is paired with an identical post-hoc
regressor $p\to z$ (MLP $2{\times}64$, 5\,000 Adam steps); JEPAWG uses its parameter path $D(P(E_p(p)))$.

\section{Intrinsic Dimension and Ablation}
\label{app:embedding}
\paragraph{Intrinsic dimension}
We estimate the intrinsic dimension of each training-set latent
with the TwoNN maximum-likelihood estimator of
Facco et al.~\citep{facco2017estimating} and the Levina--Bickel MLE
(MacKay--Ghahramani correction)~\citep{levina2004maximum} at neighbourhood
sizes $k\in\{5,10,20\}$, probing increasingly coarse scales.

\begin{table}[ht]
\centering
\caption{Intrinsic-dimension estimates across neighbourhood scale
(TwoNN, then Levina--Bickel with $k{=}5,10,20$) for the weight grids of
four families and their embeddings (all bottleneck $d_z{=}16$; PCA:
rank 16). The JEPAWG latent dimension is closest to $\dim p$ in 15 of 16 columns (exception: TwoNN on R\"ossler); PCA, AE and VAE are fitted without access to $p$.}
\label{tab:embed}
\begin{tabular}{llcccc}
\toprule
grid & method & TwoNN & $k{=}5$ & $k{=}10$ & $k{=}20$ \\
\midrule
Lorenz & raw $\theta$ & 19.9 & 17.9 & 15.1 & 12.3 \\
($\dim p{=}1$) & PCA-16 & 4.7 & 4.1 & 3.5 & 3.3 \\
 & AE  & 7.9 & 7.3 & 6.5 & 5.7 \\
 & VAE & 11.4 & 10.4 & 9.7 & 8.9 \\
 & JEPAWG & \textbf{2.9} & \textbf{2.3} & \textbf{1.7} & \textbf{1.4} \\
\midrule
R\"ossler & raw $\theta$ & 2.8 & 8.9 & 14.2 & 17.1 \\
($\dim p{=}1$) & PCA-16 & 2.5 & 4.9 & 5.6 & 5.2 \\
 & AE  & \textbf{1.4} & 4.0 & 5.7 & 6.4 \\
 & VAE & \textbf{1.4} & 4.5 & 7.0 & 8.8 \\
 & JEPAWG & 2.8 & \textbf{2.4} & \textbf{2.0} & \textbf{1.7} \\
\midrule
Brusselator & raw $\theta$ & 5.7 & 4.6 & 4.5 & 4.5 \\
($\dim p{=}2$) & PCA-16 & 3.3 & 2.7 & 2.5 & 2.4 \\
 & AE  & 3.3 & 2.8 & 2.8 & 2.8 \\
 & VAE & 3.3 & 3.0 & 3.2 & 3.4 \\
 & JEPAWG & \textbf{2.8} & \textbf{2.4} & \textbf{2.4} & \textbf{2.3} \\
\midrule
H\'enon & raw $\theta$ & 6.0 & 5.4 & 5.6 & 6.1 \\
($\dim p{=}2$) & PCA-16 & 4.0 & 3.4 & 3.1 & 2.8 \\
 & AE  & 3.2 & 2.9 & 2.8 & 2.8 \\
 & VAE & 3.2 & 3.1 & 3.3 & 3.6 \\
 & JEPAWG & \textbf{3.0} & \textbf{2.6} & \textbf{2.4} & \textbf{2.2} \\
\bottomrule
\end{tabular}
\end{table}

On the two one-parameter grids, JEPAWG's estimate decreases
monotonically with scale ($2.9\to1.4$ on
Lorenz and $2.8\to1.7$ on R\"ossler (Table~\ref{tab:embed})); on the
two-parameter Brusselator and H\'enon grids ($d{=}2$) the same descent
levels off at $2.3$ and $2.2$ respectively. The other baselines generally show much higher intrinsic dimension, further removed from the control dimension.
The raw weight grids themselves measure $12$--$20$ (Lorenz),
up to $17$ (R\"ossler), and $4.5$--$6$ (Brusselator, H\'enon). JEPAWG's
covariance-free VICReg objective acts as a regulariser, smoothing the weight variation
that the control encoder cannot predict, collapsing to a latent space that is more closely aligned with the underlying manifold.
\paragraph{Decoded weights versus raw weights}
As an ablation, we evaluate the raw weights on the forecasting task and compare with the performance of the decoded weights.  We observe that the decoded weights outperform the raw weights in $\mathrm{TtT}$ on all four systems (Table~\ref{tab:raw-vs-decoded}), even though the decoder is trained only
to reconstruct them. A plausible explanation is that each $\theta_i$ carries noise specific
to the trajectories of its fine-tune; the smooth decoder cannot fit
this per-point noise and would then converge toward a denoised weight
sheet, so that its reconstruction residual consists largely of the
component that harms forecasting.

\begin{table}[ht]
\centering
\caption{Shared $p$-conditional base network, raw fine-tuned grid weights
and JEPAWG-decoded weights on the held-out interpolation test sets. The
base $\theta_0$ is the single conditional TCNN-CD every zoo member was
fine-tuned from; raw weights use the nearest grid point's $\theta_i$; both
are deterministic given the zoo. Decoded $\mathrm{TtT}$ entries are mean $\pm$ std over 5 hypernetwork seeds; decoded MSE entries are seed means. $\mathrm{TtT}_{0.2}$ in time units (H\'enon:
iterations); one-step MSE in the standardised training space. The decoder
has the lowest one-step error on the three continuous systems and recovers
the R\"ossler horizon the per-parameter fine-tunes lose; on H\'enon
fine-tuning gives no improvement over the base.}
\label{tab:raw-vs-decoded}
\setlength{\tabcolsep}{4pt}
\begin{tabular}{lcccccc}
\toprule
 & \multicolumn{3}{c}{$\mathrm{TtT}_{0.2}$}
 & \multicolumn{3}{c}{one-step MSE} \\
\cmidrule(lr){2-4}\cmidrule(lr){5-7}
system & base $\theta_0$ & raw $\theta_i$ & decoded $\hat\theta$
       & base $\theta_0$ & raw $\theta_i$ & decoded $\hat\theta$ \\
\midrule
Lorenz      & $\mathbf{20.64}$ & $15.79$ & $20.19 \pm 0.15$
            & $1.8\text{e-}3$ & $2.5\text{e-}3$ & $\mathbf{6.7\text{e-}4}$ \\
R\"ossler   & $42.08$ & $15.33$ & $\mathbf{47.42 \pm 0.90}$
            & $2.1\text{e-}3$ & $6.6\text{e-}4$ & $\mathbf{1.1\text{e-}4}$ \\
Brusselator & $17.44$ & $24.25$ & $\mathbf{25.40 \pm 0.51}$
            & $9.7\text{e-}4$ & $6.8\text{e-}4$ & $\mathbf{8.6\text{e-}5}$ \\
H\'enon     & $\mathbf{856}$   & $733$   & $\mathbf{772 \pm 85}$
            & $\mathbf{9.6\text{e-}6}$ & $1.3\text{e-}5$ & $1.5\text{e-}5$ \\
\bottomrule
\end{tabular}
\end{table}

\begin{figure}[t]
    \centering
    \begin{minipage}[t]{0.49\linewidth}
        \centering
        \includegraphics[width=\linewidth]{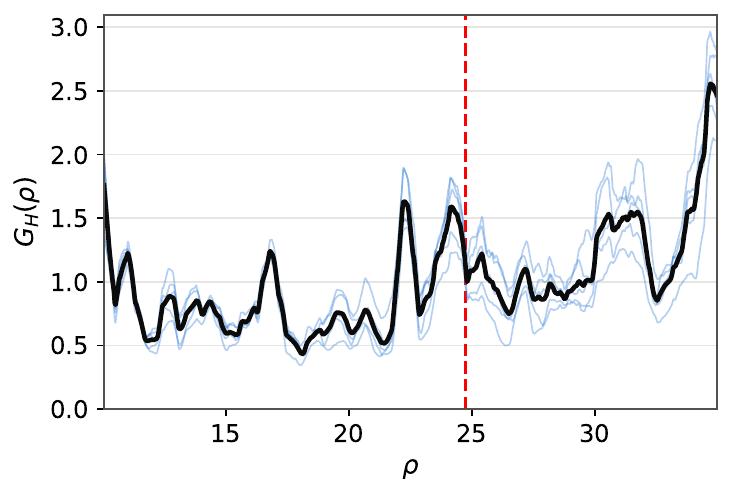}
        \caption{$G_H(\rho)$ on the Lorenz zoo for a
regressor $p \mapsto \hat\theta$ trained jointly, i.e.\ parameter encoder,
predictor and decoder fitted together on the weight-space MSE. Thin lines: five seeds; thick line: seed mean;
$G_H$ is the norm of $\mathrm{d}H(p)/\mathrm{d}\rho$; red dashed
line: subcritical Hopf point $\rho_H \approx 24.74$.}
        \label{fig:lorenzGjoint}
    \end{minipage}
    \hfill
    \begin{minipage}[t]{0.48\linewidth}
        \centering
        \includegraphics[width=0.79\linewidth]{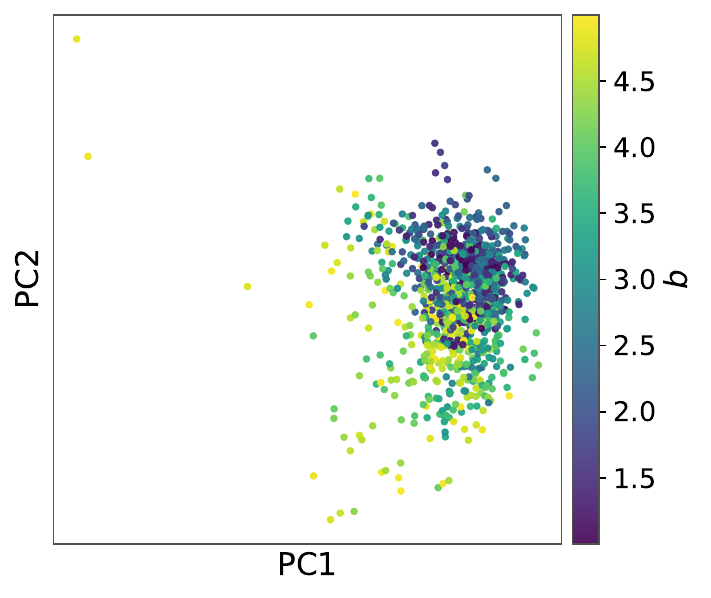}
        \caption{Weight-side latent of the Brusselator zoo without stage~1.
Each of the $M{=}1024$ fine-tuned networks $\theta_i$ is embedded as
$E_\theta(\theta_i)$ with the weight encoder and predictor left at their
random initialisation.
The first two principal components, which carry 50\,\% of the variance, are 
coloured by $b$. }

        \label{fig:brusselator-frozen}
    \end{minipage}
\end{figure}

\paragraph{Architecture ablation}
To isolate the effect of the JEPAWG architecture, we consider two alternative networks. First, we train a single network $H: p \mapsto \hat\theta$ that maps parameters directly to weight space. $H$ uses the same MLP components as JEPAWG, $H(p) = D(P(E_p(p)))$ (Sec.~\ref{sec:jepawg}), but is trained directly with the reconstruction loss on the decoded weights (Eq.~\eqref{eq:stage2}). Since no weight-space encoder $\theta \mapsto z_w$ is learned, the latent representation of $p$ is not tied to the structure of the weights, and $H$ can fit each training target independently. Fig.~\ref{fig:lorenzGjoint} shows that the resulting response $G_H(p) = \|d\hat\theta/dp\|$ is considerably noisier on the Lorenz system than under full JEPAWG training (Fig.~\ref{fig:lorenz-bif}). Although a peak is visible at $\rho_H$, comparable or larger peaks appear elsewhere, making it impossible to use $G_H$ to locate the bifurcation reliably.

As a second ablation, we freeze the encoders $E_p, E_\theta$ and the predictor $P$ (Sec.~\ref{sec:jepawg}) at initialisation and train only the decoder $D$ with the reconstruction loss (Eq.~\eqref{eq:stage2}). The frozen $E_\theta$ is then a random projection of the weights whose latent shows no organisation by parameter (Fig.~\ref{fig:brusselator-frozen}), in contrast to the trained latent of Fig.~\ref{fig:brusselator-latent}, where the same zoo lies on a surface parametrised by $(a,b)$ that separates the stable fixed point from the limit-cycle regime. The parameter encoder $E_p$ effectively becomes a random-feature embedding. On the Brusselator zoo this control reaches a
reconstruction error comparable to JEPAWG (random-feature MSE $0.0446$ vs.\ JEPAWG $0.0441$),
and its response $G$ on the $(a,b)$ grid correlates strongly with JEPAWG's
(Pearson $r = 0.96$), indicating that a similar map $p \mapsto \hat\theta$ is learned. However, a random-feature encoder fixes the geometry of
the parameter latent before any data are seen: every coordinate of $p$ is
embedded on the same footing, irrespective of its effect on the dynamics. JEPAWG's latent is instead shaped by the zoo through the invariance term,
which ties $P(E_p(p))$ to the weight-side latent $E_\theta(\theta)$. A parameter that moves the weights spans a latent direction, while one that does not is dropped. With a single parameter, the two embeddings behave alike, since there is
only one direction along which to allocate capacity. They separate as soon as the
coordinates of $p$ are not on equal footing, which is the common case in more complicated
systems with several parameters of unequal influence, or with parameters that
are recorded but do not enter the dynamics.

\begin{figure}[t]
    \centering
    \begin{subfigure}[t]{0.49\linewidth}
        \centering
        \includegraphics[width=\linewidth]{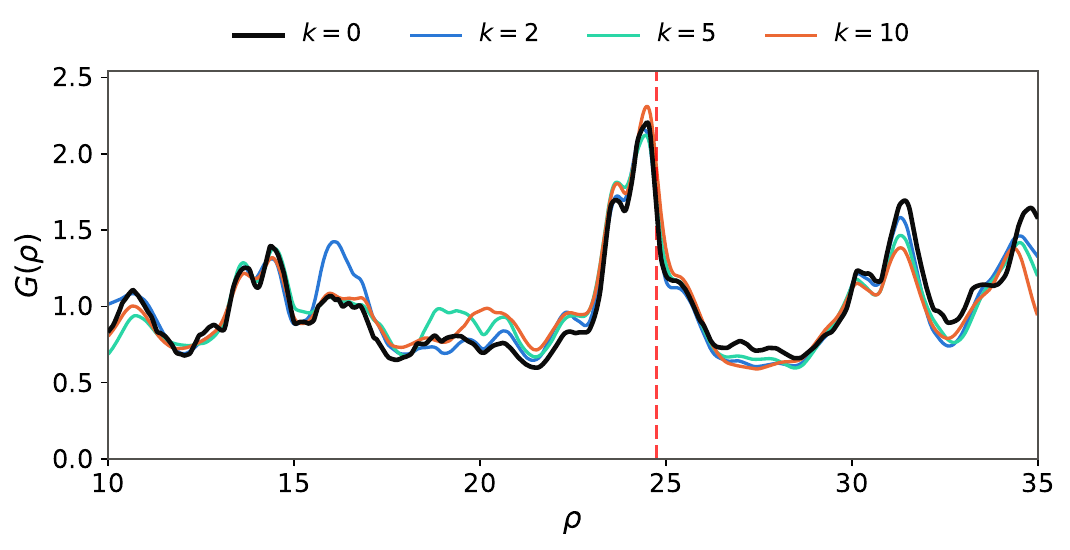}
        \caption{JEPAWG encoding.}
        \label{fig:randomp_jepawg}
    \end{subfigure}
    \hfill
    \begin{subfigure}[t]{0.49\linewidth}
        \centering
        \includegraphics[width=\linewidth]{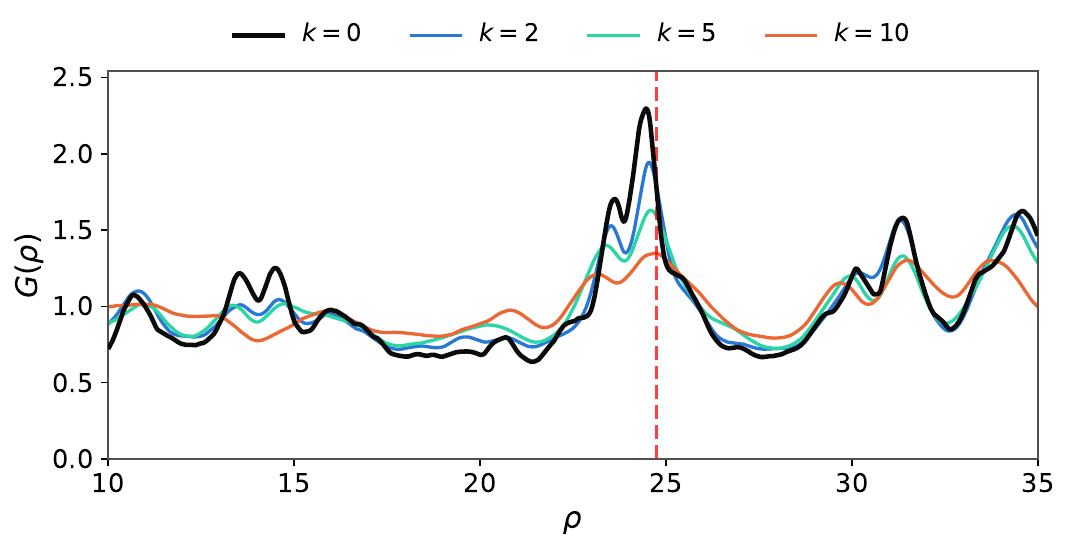}
        \caption{Random-feature encoding.}
        \label{fig:randomp_frozen}
    \end{subfigure}
    \caption{\textbf{Random parameters.} $k$ random coordinates
$\xi \sim \mathcal{U}(0,1)^k$, with no effect on the dynamics, are appended
to $p = (\rho, \xi)$ on the Lorenz zoo ($M{=}800$, weights unchanged), and
both models are retrained. Shown is $G(\rho)$ evaluated at a fixed random
draw of $\xi$, averaged over 5 seeds; the red line marks
$\rho_H \approx 24.74$. (a) JEPAWG; the curves for $k = 0, 2, 5, 10$
coincide. (b) the random-feature control, with the same architecture
but encoders frozen at initialisation. As $k$ grows, its baseline rises,
the Hopf peak broadens and loses prominence.}
    \label{fig:randomp}
\end{figure}

We test the limiting case directly by appending $k$ random parameters
$\xi \sim \mathcal{U}(0,1)^k$ to $p = (\rho, \xi)$ on the Lorenz zoo; they
have no effect on the dynamics, and the zoo weights are unchanged. JEPAWG is insensitive to $\xi$: the Hopf peak keeps its location (Fig.~\ref{fig:randomp}), and at most $0.1\%$ of the squared gradient of $\hat\theta$
lies along $\xi$. The random-feature control has no way to tell $\xi$ apart
from $\rho$ and spends capacity on it, so its peak near the Hopf point
degrades as $k$ grows, until $G$ no longer reliably locates the
bifurcation. The forecasts reflect the same difference. At $k = 10$, the interpolation
horizon of the JEPAWG-decoded networks is $20.1 \pm 0.2$ time units,
against $20.2 \pm 0.2$ at $k = 0$, while that of the random-feature control
drops from $20.2 \pm 0.2$ to $18.5 \pm 0.5$.

\section{Extrapolation per Regime}
\begin{figure}[t]
    \centering
    \includegraphics[width=\linewidth]{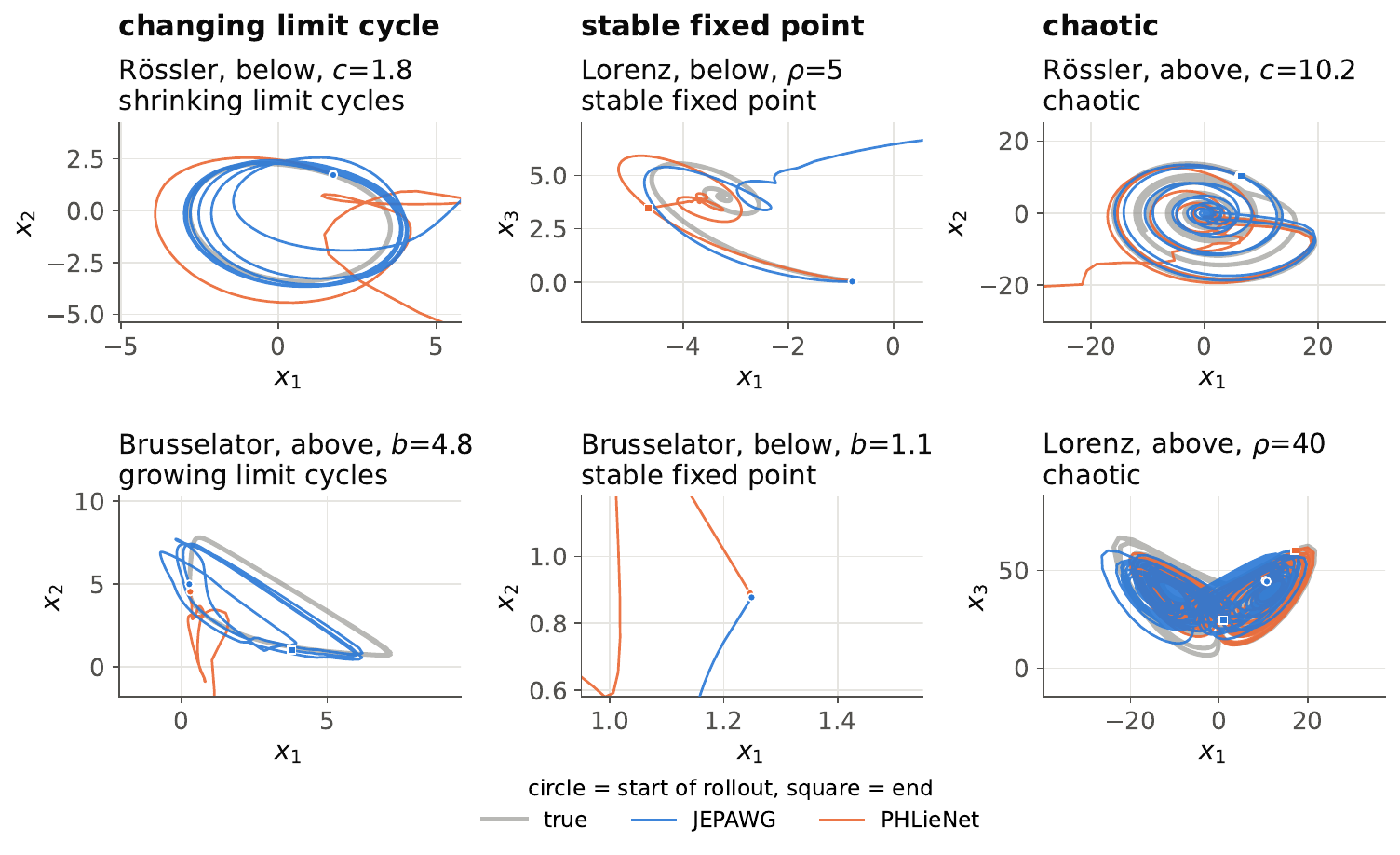}
\caption{Predicted and true trajectories at the furthest extrapolation
parameter in each regime (one seed). In the limit-cycle regimes,
JEPAWG-generated forecasters follow the true cycle more closely, while
PHLieNet-generated trajectories
tend to diverge after a number of steps, as quantified by the bounded
fraction in Table~\ref{tab:extrap_regimes}.}
    \label{fig:extrap_regimes_deepest}
\end{figure}

Figure~\ref{fig:extrap_regimes_deepest} shows predicted and true trajectories at the furthest extrapolation parameter of each regime in Table~\ref{tab:extrap_regimes}.

\section{Additional Results per Dynamical System}
\subsection{Lorenz}
On the Lorenz forecasting experiment, we plot JEPAWG forecasts and validate generated trajectories against the numerically integrated evolution.

\begin{figure}[H]
    \centering
    \includegraphics[width=0.9\linewidth]{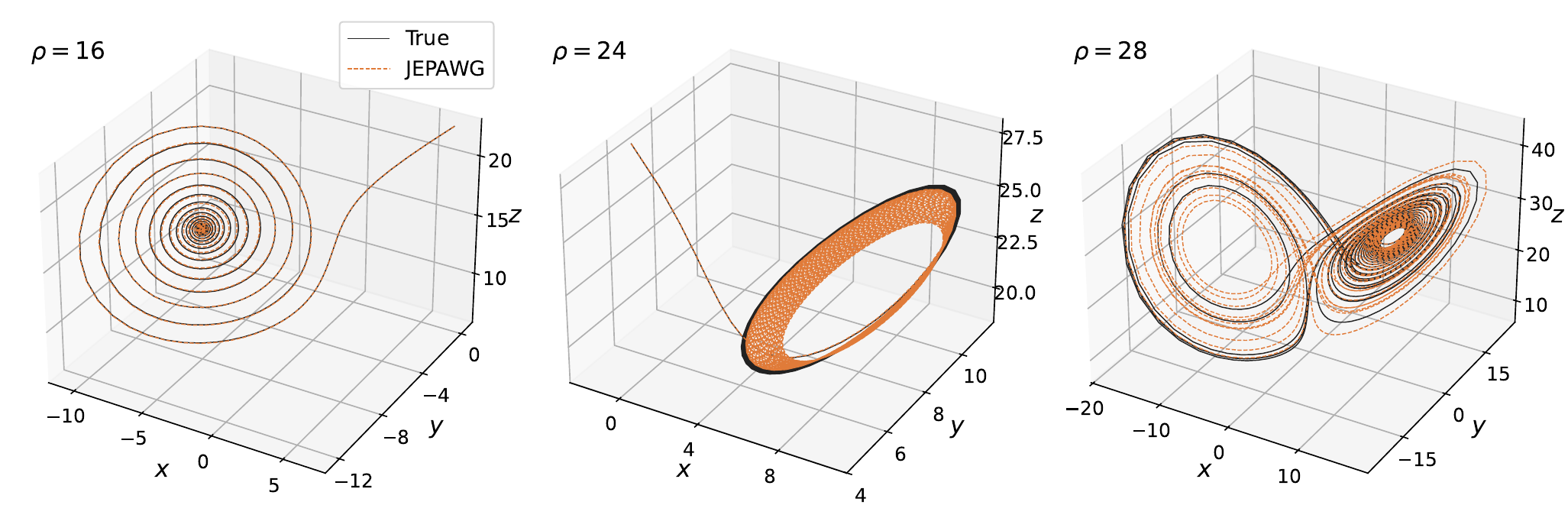}
    \caption{Forecast trajectories of the Lorenz system at held-out $\rho$ (solid: truth;
dashed: JEPAWG) at $\sigma = 10$. From left to right: spiral trajectory into a stable fixed point, slowly decaying spiral just below the subcritical Hopf point, and the chaotic
attractor. At these $\rho$ inside the training range, the forecasts follow the true trajectory with reasonable precision.} \label{fig:lorenz-rollouts}
\end{figure}
\subsection{R\"ossler}
\label{app:rossler}
For the R\"ossler system we also examine $G(c)$ over the range $c\in [3,9]$, and observe distinct peaks in the response at period doublings and transitions into chaos.

\begin{figure}[H]
    \centering
    \includegraphics[width=0.65\linewidth]{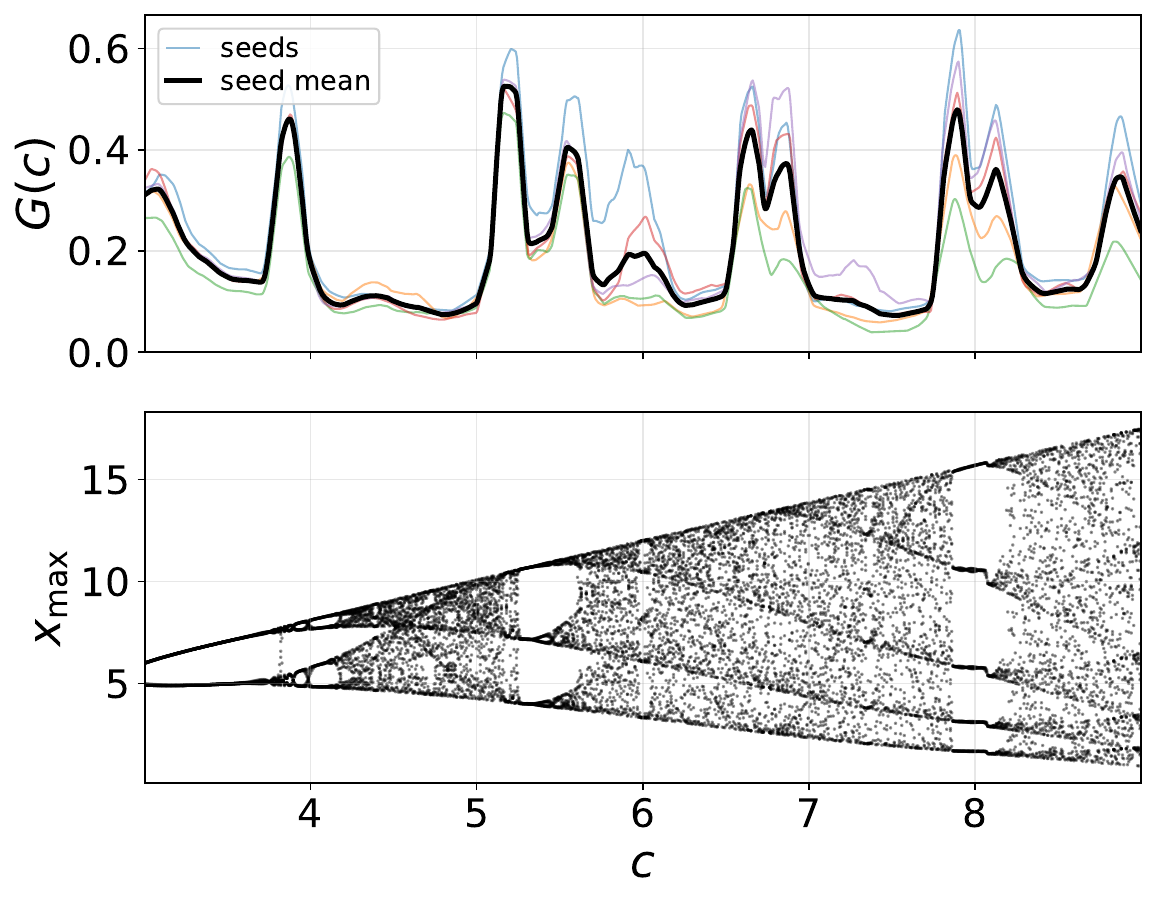}
    \caption{R\"ossler weight-space response at $M{=}1000$ ($a{=}b{=}0.2$,
$c\in[3,9]$). Top: the response $G(c)$ of JEPAWG-generated weights (Gaussian smoothed), per-seed curves and seed
mean. Bottom: ground-truth diagram (successive local maxima of $x(t)$
per parameter). The peaks of $G$ correspond to features of the bifurcation
diagram in the plot below: the $2\!\to\!4$ doubling at
$c{=}3.83$ and the entry/exit boundaries of the periodic windows show
clear signal; the finite parameter resolution limits the visibility of
the doublings closer to chaos.} \label{fig:rossler-G}
\end{figure}
\subsection{H\'enon}
\label{app:henon}
\begin{figure}[H]
    \centering
    \includegraphics[width=0.8\linewidth]{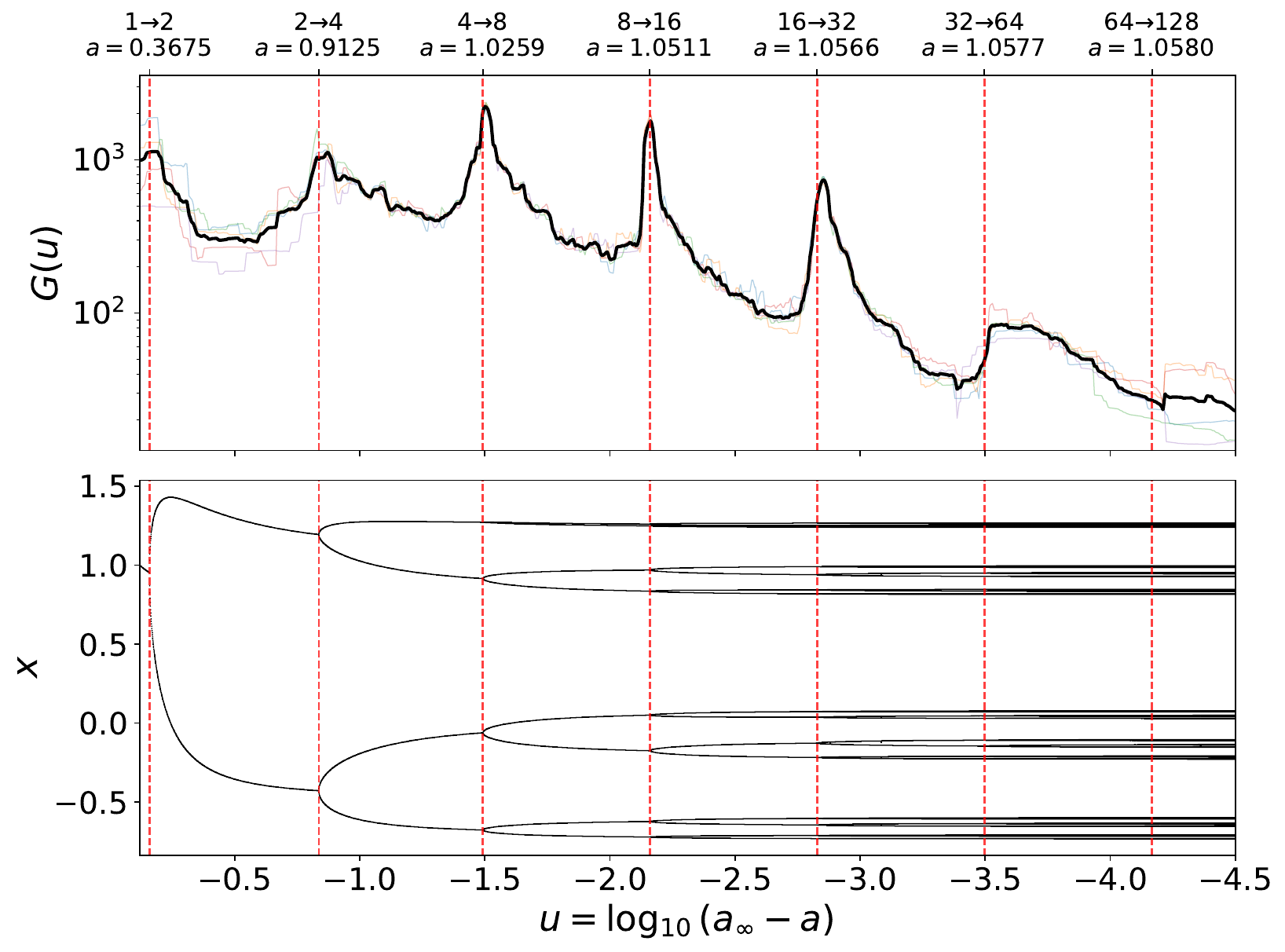}
\caption{H\'enon $u$-cascade. Top: weight-space response
$G(u)$ of the cascade zoo, 5 seeds in colour and their mean in black,
over the full trained range $u\in[-4.5,-0.12]$ (Fig.~\ref{fig:henon-ucascade}
shows the part $u\ge-3.2$). Bottom: bifurcation diagram of the H\'enon
map ($b{=}0.3$) in the same coordinate $u=\log_{10}(a_\infty-a)$, with
$a_\infty = 1.05805$ \cite{collet1981}. For each of $6000$ parameter
values, $512$ iterates of the attractor are shown after a
$2{\times}10^5$-iterate transient. Red dashed lines mark the
period-doubling bifurcations $a_1,\dots,a_7$ (labels on the top axis);
$a_1$ is exact, $a_2,\dots,a_7$ were located by bisection on the attractor
period. The last two
splittings of the orbit points are narrower than $10^{-3}$ in $x$ and are not resolved at this
scale. $G(u)$ peaks at each of the first five doublings, shows a step at
$a_6$ and no feature at $a_7$, where the seeds begin to disagree.}
    \label{fig:henon-perioddoubling}
\end{figure}
\begin{figure}[H]
\centering
\includegraphics[width=0.98\textwidth]{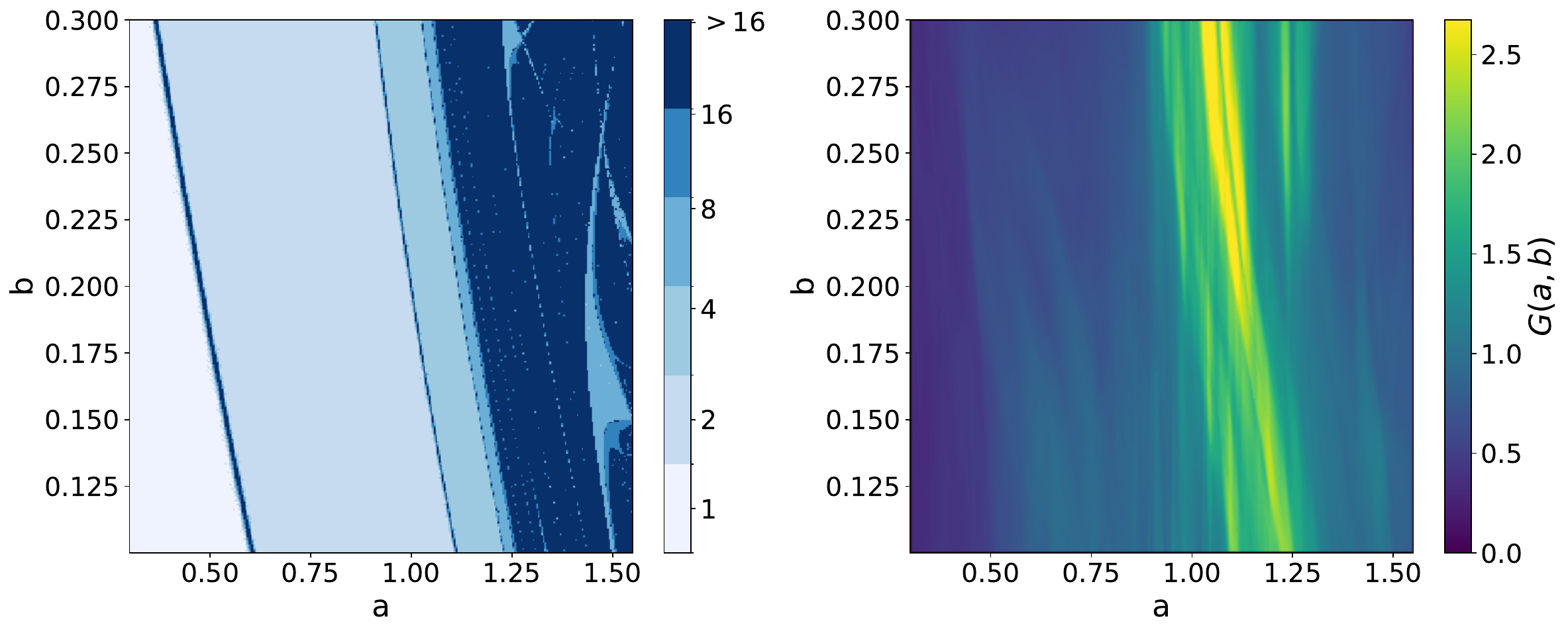}
\caption{H\'enon. Left: ground-truth orbit period over $(a,b)$; the
boundaries between coloured bands are the bifurcation curves. Right:
JEPAWG weight-space response $G(a,b)=\|\nabla_{a,b}\hat{\theta}\|$ on the same rectangle, mean over
seeds. The ridge of $G$ tracks the interior doubling curves and the
chaotic band.}
\label{fig:henon-period}
\end{figure}

\paragraph{Two-parameter grid}
Trained on the 2-D grid $(a,b)\in[0.3,1.55]\times[0.1,0.3]$
($M{=}1024$), the weight-space response reproduces the interior structure of
the ground-truth period map (Fig.~\ref{fig:henon-period}): the ridge of
$\|\nabla_{a,b}\hat{\theta}\|$ follows the $2\!\to\!4$ and $4\!\to\!8$
doubling curves and broadens over the chaotic band.

\section{Compute Resources}
\label{app:compute}
All training and evaluation ran on single NVIDIA L4 GPUs on a
university SLURM cluster. The dominant cost is the construction of each weight grid:
conditional pre-training followed by $M$ per-parameter fine-tunes
takes 2--6 GPU-hours per grid, depending on trajectory length and
grid size. Training one JEPAWG seed on a finished grid takes
$\sim$40 seconds, so the five-seed embedding stage is negligible
relative to grid construction; evaluation adds 5--20 GPU-minutes
per system.
\section{Experimental Details}
\begin{table}[H]
\centering
\caption{Per-experiment protocol. The $M$ parameter values are drawn
from scrambled Sobol sequences (seed 0; stratified per-band draws for
the Brusselator $b$ sweep and the H\'enon $u$-cascade, joint draws
for the two-parameter families). Transients are integrated \emph{in
addition to} the recorded lengths; a zero entry means transients are
retained in the data. $\Delta t$: observation interval (maps: 1 iteration);
recorded lengths as training/test. ICs: initial conditions per
training parameter (test sets use 10; Lorenz 20). $h$: evaluation
step of the $G$ sweep (two-parameter rows list the per-axis pair
$(h_1, h_2)$ in the order of the experiment label), alongside the mean spacing $\Delta p$ of the fine-tuned parameters (Brusselator $b$: spacing in the dense band $b\in[2.35,2.85]$ around $b_c$; overall mean $2.4{\times}10^{-3}$) ($u$-cascade:
both in $u=\log_{10}(a_\infty-a)$, on which it is trained).
$\lambda_{\max}^{\sup}$: largest Lyapunov exponent over the training
range (per time unit; maps: per iteration); $\le 0$ marks ranges
without chaotic parameters. Input
noise $5\%$ of the standardised feature scale. The Lorenz-$\rho$ grid also underlies Fig.~\ref{fig:lorenz-bif}.}
\label{tab:protocol}
\footnotesize
\setlength{\tabcolsep}{3pt}
\begin{tabular}{lcccccccc}
\toprule
exp. & $M$ & trans. & $\Delta t$ & len (tr/te) & ICs & $h$ & $\Delta p$ & $\lambda_{\max}^{\sup}$ \\
\midrule
\multicolumn{9}{l}{\emph{Forecasting}} \\
Lorenz, $\rho$          & 800  & 0       & 0.02 & 20/30     & 5  & $3.1{\times}10^{-2}$ & $3.1{\times}10^{-2}$ & 1.06 \\
R\"ossler, $c$          & 1000 & 300     & 0.1  & 20/100    & 10 & $7.5{\times}10^{-3}$ & $6.0{\times}10^{-3}$ & 0.09 \\
Brusselator, $b$        & 1024 & 20      & 0.05 & 30/30     & 5  & $3.1{\times}10^{-3}$ & $8{\times}10^{-4}$ & $\le 0$ \\
H\'enon, $a$     & 1024 & 200 it. & 1    & 800/1500  & 5  & $3.1{\times}10^{-4}$ & $2.4{\times}10^{-4}$ & 0.18 \\
\midrule
\multicolumn{9}{l}{\emph{Interpretability}} \\
Lorenz, $(\sigma,\rho)$ & 1024 & 0 & 0.02 & 20/30 & 5 & $(0.75,\, 1.5){\times}10^{-1}$ & $6.6{\times}10^{-1}$ & 1.14 \\
Lorenz, T-point $(\sigma,\rho)$ & 1024 & 0 & 0.02 & 20/30 & 5 & $(1.5,\, 2.1){\times}10^{-2}$ & $1.3{\times}10^{-1}$ & 1.03 \\
Brusselator, $(a,b)$ & 1024 & 20 & 0.05 & 30/30 & 5 & $(0.75,\, 2.0){\times}10^{-2}$ & $7.7{\times}10^{-2}$ & $\le 0$ \\
H\'enon, $(a,b)$ & 1024 & 200 it. & 1 & 800/1500 & 5 & $(4.2,\, 0.67){\times}10^{-3}$ & $1.6{\times}10^{-2}$ & 0.45 \\
H\'enon, $u$ & 1540 & 0 & 1 & 1600/1500 & 5 & $3.7{\times}10^{-4}$ & $2.8{\times}10^{-3}$ & $\le 0$ \\
\bottomrule
\end{tabular}
\end{table}

\begin{table}[H]
\centering
\caption{Interpolation and extrapolation test sets for the forecasting
experiments. Interpolation test parameters are Sobol draws from the training
range, except Lorenz-$\rho$, which uses 10 fixed values. Fixed parameters:
Lorenz $\sigma{=}10$, $\beta{=}8/3$; R\"ossler $a{=}b{=}0.2$; Brusselator $a{=}1.25$; H\'enon $b{=}0.3$.}
\label{tab:testsets}
\small
\setlength{\tabcolsep}{4pt}
\begin{tabular}{@{}llll@{}}
\toprule
experiment & train range & interp.\ test & extrapolation \\
\midrule
Lorenz, $\rho$   & $[10,35]$      & 10 fixed & $36$--$40$; low $5$--$9$ \\
R\"ossler, $c$   & $[3,9]$        & 10 Sobol & $1.8$--$2.8$; $9.2$--$10.2$ \\
Brusselator, $b$ & $[1.5,4]$ & 32 Sobol & $1.1$--$1.4$; $4.2$--$4.8$ \\
H\'enon, $a$ & $[0.85,1.10]$ & 32 Sobol & --- \\
\bottomrule
\end{tabular}
\end{table}

\end{document}